\documentclass[10pt,journal,compsoc]{IEEEtran}

\usepackage{xcolor}
\usepackage{booktabs}
\usepackage{epsfig}
\usepackage{graphicx}
\usepackage{amsmath}

\usepackage{amssymb}

\usepackage{verbatim}
\usepackage{changepage}
\usepackage{url}
\usepackage{tabularx}
\usepackage{multirow}
\usepackage{subfigure}
\usepackage{amsmath,dsfont}
\usepackage{array}
\usepackage{enumitem}
\usepackage[utf8]{inputenc}
\usepackage{pifont}
\usepackage{threeparttable}

\usepackage{verbatim}

\usepackage{colortbl}
\definecolor{mygray}{gray}{.75}

\usepackage{xcolor}
\usepackage{bbding}
\usepackage{xspace}

\usepackage[normalem]{ulem}

\newcommand{\eg}{\emph{e.g.,}~}
\newcommand{\etal}{\emph{et al.}~}
\newcommand{\ie}{\emph{i.e.,}~}
\newcommand{\wrt}{\emph{w.r.t.}\xspace}

\ifCLASSOPTIONcompsoc
  \usepackage[nocompress]{cite}
\else
  \usepackage{cite}
\fi
\ifCLASSINFOpdf
\else
\fi
\begin{document}
%
\title{PRVR: Partially Relevant Video Retrieval}
%
%
%
%

\author{Xianke~Chen, Daizong~Liu,  Xun~Yang,
        Xirong~Li,~\IEEEmembership{Member,~IEEE,}
         Jianfeng~Dong, \\
        Meng~Wang,~\IEEEmembership{Fellow,~IEEE,}
        Xun~Wang,~\IEEEmembership{Member,~IEEE}
\IEEEcompsocitemizethanks{\IEEEcompsocthanksitem X. Chen is with the School of Computer Science and Technology, and the School of Statistics and Mathematics, Zhejiang Gongshang University, Hangzhou 310035, China. E-mail: cxkxk\_@outlook.com \protect
\IEEEcompsocthanksitem J. Dong and X. Wang are with the the School of Computer Science and Technology, Zhejiang Gongshang University, and the Zhejiang Key Laboratory of Big Data and Future E-Commerce Technology,  Hangzhou 310035, China. 
\protect
E-mail: dongjf24@gmail.com, wx@zjgsu.edu.cn 

\IEEEcompsocthanksitem D. Liu is with the Wangxuan Institute of Computer Technology, Peking University, No. 128, Zhongguancun North Street, Beijing 100871, China. 
\protect
E-mail: dzliu@stu.pku.edu.cn

\IEEEcompsocthanksitem X. Yang is with the School of Information Science and Technology, University of Science and Technology of China, Hefei 230026, China.
\protect
E-mail: xyang21@ustc.edu.cn

\IEEEcompsocthanksitem X. Li is with the School of Information, Renmin University of China, Beijing 100872, China.
\protect
E-mail: xirong@ruc.edu.cn

\IEEEcompsocthanksitem M. Wang is with the School of Computer Science and Information
Engineering, Hefei University of Technology, Hefei 230009, China.
\protect
E-mail: wangmeng@hfut.edu.cn}
\thanks{Manuscript received 15 Oct, 2024; revised 15 Jul, 2025 and 4 Sep, 2025; accepted Sep 18, 2025. \\
(Corresponding authors: Jianfeng Dong, Xirong Li and Xun Wang)}}

%
%

\markboth{IEEE Transactions on Pattern Analysis and Machine Intelligence,~Vol.~X, No.~X, September~2025}%
{Shell \MakeLowercase{\textit{et al.}}: Partially Relevant Video Retrieval}
%



\IEEEtitleabstractindextext{%
\begin{abstract}

In current text-to-video retrieval (T2VR), videos to be retrieved have been properly trimmed so that a correspondence between the videos and ad-hoc textual queries naturally exists. Note in practice that videos circulated on the Internet and social media platforms, while being relatively short, are typically rich in their content. Often, multiple scenes / actions / events are shown in a single video, leading to a more challenging T2VR setting wherein only part of the video content is relevant w.r.t. a given query. This paper presents a first study on this setting which we term Partially Relevant Video Retrieval (PRVR).
Considering that a video typically consists of multiple moments, a video is regarded as \emph{partially} relevant w.r.t. to a given query if it contains a query-related moment. 
We formulate the  PRVR task as a multiple instance learning problem, and propose a Multi-Scale Similarity Learning (MS-SL++) network that jointly learns both clip-scale and frame-scale similarities to determine the partial relevance between video-query pairs. Extensive experiments on three diverse video-text datasets (TVshow Retrieval, ActivityNet-Captions and Charades-STA) demonstrate the viability of the proposed method. Source code and datasets are available at \url{https://github.com/HuiGuanLab/ms-sl-pp}.
\end{abstract}

\begin{IEEEkeywords}
Video-Text Retrieval, Cross-Modal Retrieval, Partially Relevant, Video Representation Learning.
\end{IEEEkeywords}}

\maketitle

\IEEEdisplaynontitleabstractindextext

%
\IEEEpeerreviewmaketitle

\begin{figure}[]
\centering
\subfigure[Current Text-to-Video Retrieval (T2VR)]{
\includegraphics[width=0.95\columnwidth]{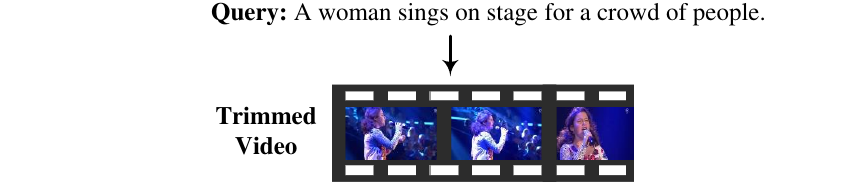}
} 
\subfigure[Proposed Partially Relevant Video Retrieval (PRVR)]{
\includegraphics[width=0.95\columnwidth]{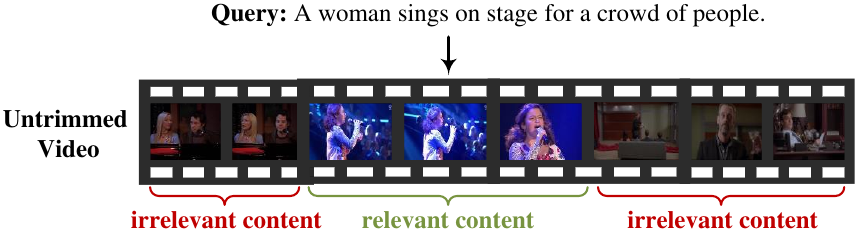}
}
\caption{Difference between current T2VR task and our proposed PRVR task. In T2VR, a target video is typically pre-trimmed and almost \textit{fully relevant} to the corresponding query, which is too idealized compared to real-world retrieval scenarios. By contrast, a target video in PRVR is untrimmed and diverse with query-irrelevant content, which is regarded as \textit{partially relevant} to the query.}\label{fig:task_diff}
\vspace{-7mm}
\end{figure}

\ifCLASSOPTIONcompsoc
\IEEEraisesectionheading{\section{Introduction}\label{sec:introduction}}
\else
\section{Introduction}
\label{sec:introduction}
\fi
\IEEEPARstart{T}{ext}-to-video retrieval (T2VR) \cite{li2020sea,sigir2020tree,croitoru2021teachtext} is a popular task in the multimedia field.
Although significant progress has been achieved in the past few years, current methods for T2VR  \cite{liu2019use,jin2021hierarchical,chen2020fine,gabeur2020multi,han2021fine,luo2022clip4clip} assume that the video to be retrieved is fully relevant \textit{w.r.t.} a given textual query. 
Moreover, they simply make the evaluations on video-captioning oriented datasets such as MSVD \cite{chen2011collecting}, MSR-VTT \cite{xu2016msr} and VATEX \cite{wang2019vatex}. 
A key property of these datasets is that videos are assumed to be temporally pre-trimmed with short duration, whilst the provided captions well describe the gist of the video content.
Consequently, for a given paired video and caption, the video is supposed to be fully relevant to the caption, as shown in Fig.\ref{fig:task_diff} (a). In reality, however, as queries are not known a priori, pre-trimmed video clips may not contain sufficient content to fully meet the query. This suggests a gap between the literature and the real world.

To fill out this gap, we propose a novel subtask of T2VR termed \textit{Partially Relevant Video Retrieval (PRVR)}, which aims to retrieve such partially relevant videos from a large collection of untrimmed videos.
As shown in Fig.\ref{fig:task_diff} (b), an untrimmed video is considered partially relevant \textit{w.r.t.} a given textual query as long as it contains a (short) moment that is relevant to the query. 
The video may include content that is irrelevant to the query.
Since both the location of the relevant moment and its duration are unknown, and the presence of irrelevant content in the target videos, PRVR is more challenging than the current T2VR task.

\begin{figure}[tb!]
\centering\includegraphics[width=0.98\columnwidth]{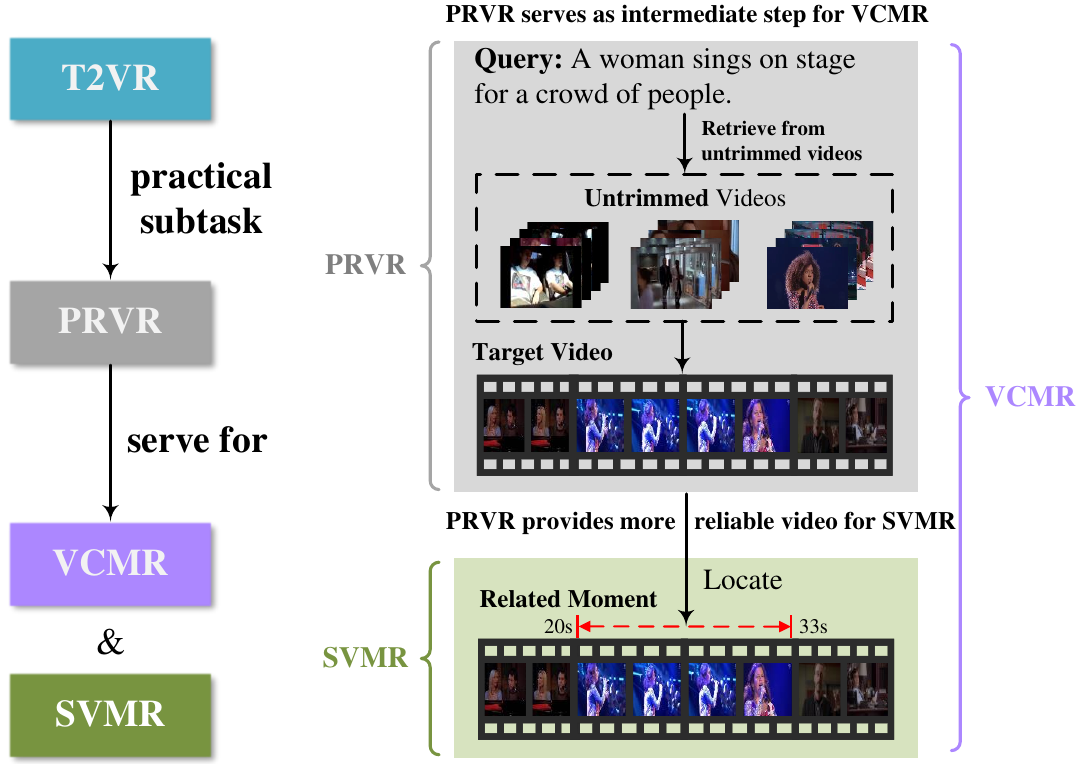}
\caption{Illustration of the connection of the proposed PRVR task with existing video/moment retrieval tasks. In particular, PRVR can be regarded as a more practical but challenging subtask of T2VR, and serves as a crucial intermediate step for VCMR and SVMR tasks by providing videos partially related to the query}.\label{fig:task_relation}
\vspace{-7mm}
\end{figure}

Additionally, it is worth noting that our proposed PRVR also serves as a crucial intermediate step for various downstream vision-language tasks, such as Single Video Moment Retrieval (SVMR)\cite{yuan2019find,zhang2020learning,xiao2021boundary,yang2021deconfounded} and Video Corpus Moment Retrieval (VCMR)\cite{lei2020tvr,zhang2020hierarchical,zhang2021video,hou2021conquer}.
Specifically, SVMR aims to identify the specific moment in an untrimmed video that semantically aligns with a given natural language query, while VCMR involves seeking out a candidate video from a video collection and then pinpointing a moment within that video based on the query.
Both tasks require fine-grained text-guided moment retrieval within a video, and our PRVR task happens to provide reliable prior steps to retrieve video partially related to the query from a large untrimmed video set for laying the foundation.
As shown in Fig.~\ref{fig:task_relation}, as a more practical subtask of T2VR, our proposed PRVR not only aids in retrieving more accurate, partially related videos during the first stage of VCMR  but also serves as an initial step for the SVMR setting.
Therefore, exploring the new PRVR task is valuable for advancing these vision-language tasks.

A model for PRVR has to be trained on partially relevant video-text pairs. In fact, training on a ``high-quality'' dataset where video-text are fully relevant is problematic, due to the significant disparity between the training data and the test data.
Therefore, we resort to three datasets, \ie TVR \cite{lei2020tvr}, ActivityNet-Captions \cite{krishna2017dense}, and Charades-STA \cite{gao2017tall}, which are constructed for SVMR. 
These datasets are appropriate for PRVR since they contain untrimmed videos with specific temporal annotations aligned with textual queries, reflecting the partially relevant nature.

To tackle this practical yet challenging PRVR task, in this paper,
we depart from the global matching paradigm used in current T2VR methods. Instead, we introduce a local matching PRVR method with multiple instance learning (MIL) \cite{dietterich1997solving,maron1997framework} framework, where a video is viewed as a bag of video clips and a bag of video frames simultaneously and we only need to find whether it has a text-related frame/clip to determine if a video is partially related to the text.
Specifically, we represent each video at both clip and frame scales to effectively handle moments of varying temporal lengths. 
Based on these representations, we introduce a Multi-Scale Similarity Learning (MS-SL++) network containing a clip-scale similarity learning branch and a frame-scale similarity learning branch, which aims to assess the partial relevance at both coarse and fine-grained temporal granularity respectively. 
The clip-scale branch is designed to identify key clips that partially overlap with the relevant moments, providing a coarse but meaningful correlation between video segments and the query. Meanwhile, the frame-scale branch enhances this by capturing fine-grained correlations at the frame level, compensating for any missed information in the clips. 
Finally, the similarities of both clip-scale and frame-scale are jointly used to measure the overall partial video-text similarity.

In sum, our main contributions are as follows: 
\begin{itemize}
    \item We propose a new T2VR subtask named PRVR, where an untrimmed video is considered to be partially relevant with respect to a given textual query if it contains a moment relevant to the query. PRVR aims to retrieve such partially relevant videos from a large collection of untrimmed videos. 
   \item  We formulate the PRVR task as a MIL problem, simultaneously viewing a video as a bag of video clips and a bag of video frames. Clips and frames represent video content at different temporal scales. Based on multi-scale video representation, we propose MS-SL++ to compute the relevance between videos and queries in a coarse-to-fine manner. 
   \item  We propose a length-aware clustering module that is able to select the representative clips from video clips obtained by the sliding window method, which can save the computation and memory costs of our model during the inference stage.
   \item  Extensive experiments on three datasets, \ie TVR~\cite{lei2020tvr}, ActivityNet-Captions \cite{krishna2017dense} and Charades-STA \cite{gao2017tall}, demonstrate the viability of the proposed method for PRVR. We also show that our method can be used for improving video corpus moment retrieval.
\end{itemize}

A preliminary version of this work was published at ACM Multimedia 2022 \cite{dong2022partially} (Oral presentation). The journal extension improves over the conference paper mainly in three aspects. 
(1) Theoretically, we provide an in-depth analysis of the necessity of the proposed PRVR task by comparing it with relevant T2VR, SVMR, and VCMR tasks. Our PRVR task is more practical and challenging than current T2VR, and serves as a crucial intermediate step for SVMR and VCMR.
(2) Technically, we enhance the original clip-scale similarity learning by introducing a length-aware clustering module, which filters redundant clip-scale representations based on temporal duration information. Compared to the previous model in~\cite{dong2022partially}, the improved MS-SL++ model achieves marginally better performance, yet much faster retrieval time and much less storage. Specifically, on the TVR dataset, it improves retrieval performance in terms of SumR from 172.4 to 174.4, reduces retrieval time by over 60\% (from 33ms to 12ms per query), and decreases feature storage by about 75\% (from 4.1GB to 1.02GB for all candidate videos), making it more suitable for practical application deployment.
(3) Experimentally, we strengthen our evaluation by including ablation analysis of the clustering module, studies on the granularity of clips, and comparison to more recent works and baseline models.
We provide a performance comparison using features extracted by the Contrastive Language-Image Pre-Training (CLIP)~\cite{radford2021learning} model. We further analyze the robustness of our model against artificial noises, the generalization ability on unseen data, and add more qualitative results.

\section{Related Work} \label{sec:rel-work}

\subsection{Text-to-Video Retrieval (T2VR)}
The T2VR task has gained much attention in recent years \cite{dong2018predicting,song2019Polysemous,chen2020fine,jin2020deep,sigir2020tree,han2021fine,liu2021progressive,wang2020learning,wu2020interpretable}, aiming to retrieve relevant videos by a given query from a set of pre-trimmed video clips. The retrieved clips are supposed to be fully relevant to the given query.
A common solution for T2VR is to first encode videos and textual queries and then map them into common embedding spaces where the cross-modal similarity is measured.
Therefore, current works mainly focus on video representation \cite{liu2019use,jin2021hierarchical,feng2021exploiting,song2021spatial,gao2020learning,gao2023vectorized,liu2023cross}, sentence representation \cite{chen2020fine,li2020sea,croitoru2021teachtext}, and their cross-modal similarity learning \cite{yu2018joint,dong2021dual,gabeur2020multi,wu2021hanet,liu2022entity}.
As our work mainly focuses on video representation and cross-modal similarity learning, we review recent progress in these two aspects in the following.

For video representation, due to the short duration of videos in conventional T2VR datasets such as MSR-VTT~\cite{xu2016msr} and VATEX~\cite{wang2019vatex}, a video is typically represented as one or several holistic feature vectors~\cite{dong2021dual,liu2019use,gabeur2020multi,eccv2022-laff}.
The common way of video representation is to first extract frame-level features by pre-trained CNN models, and then aggregate them by mean pooling~\cite{dong2018predicting,mithun2018learning} or max pooling~\cite{miech2019howto100m,miech2018learning} to obtain a video-level feature.
Since the simple pooling strategies are insufficient to summarize complete information about the whole video, more complicated methods are proposed. Dong \etal \cite{dong2019dual} jointly utilize mean pooling, bi-GRU and biGRU-CNN to obtain three-level feature vectors and concatenate them as the final video representation. 
In \cite{liu2019use}, Liu \etal employ multiple pre-trained experts to extract multi-modal features such as scene, actions audio and objects, and fuse them with a gating attention mechanism into a feature vector.
Recently, a large-scale image-language pre-training model CLIP\cite{radford2021learning} is wildly used in T2VR task and brings significant performance gains~\cite{bain2021frozen,luo2022clip4clip,fang2021clip2video,max2022guide,zhao2022centerclip,wu2023cap4video,tian2024towards,wang2024text}. In these works, patch-level or frame-level features are usually extracted by CLIP, and are subsequently aggregated into a video-level feature.
For instance, Luo \etal \cite{luo2022clip4clip} respectively try mean pooling, Transformer Encoder, and LSTM to aggregate frame-level features, while Max \etal \cite{max2022guide} use the frame-wise attentions. In \cite{wang2024text}, patch-level features are aggregated into a vector after interactive encoding with text query.
Additionally, we observe an increasing trend of representing video in several feature vector~\cite{wray2019fine,chen2020fine,dong2022reading,eccv2022-laff,zhang2023multi}. 
For instance, Chen \etal \cite{chen2020fine} use three independent transformation matrices to encode videos into three-level features, \ie the global event level, local action and entity level.
In \cite{dong2022reading}, Dong \etal represent videos into two features of various granularities by a previewing branch and an intensive-reading branch.
Zhang \etal \cite{zhang2023multi} utilize the clustering method to select a set of frame-level features to represent key events in the video.

For cross-modal similarity learning, existing T2VR methods mainly focus on the holistic similarity between videos and text~\cite{li2019w2vv++,dong2019dual,chen2020fine,wang2021t2vlad,shvetsova2022everything}. 
The most popular way is to learn a common space for videos and text, and the cross-modal similarity is measured by a standard similarity metric between two feature vectors, \eg cosine similarity~\cite{li2019w2vv++,dong2019dual,cao2022visual,gorti2022x,liu2022ts2}. Recently, instead of a common space, learning multiple common spaces has achieved increasing attention~\cite{dong2022reading,li2020sea,dong2021dual,bai2022lat,fang2022concept}.
After being projected in multiple common spaces, the final video-text similarity is computed as the weighted sum of their similarities in the multiple spaces.
For instance, Wray \etal \cite{wray2019fine} learning two different latent spaces corresponding to nouns and non-noun words respectively. In \cite{dong2021dual}, a latent space and a concept space are jointly learned for similarity prediction.

Different from the above works, we consider a more realistic scenario, 
where videos are supposed to be partially relevant to a specific query. We thus focus more on how to measure the partial relevance between text and videos.

\subsection{Video Moment Retrieval (VMR)}
The VMR task is to retrieve moments semantically relevant to the given query from a given single untrimmed video or a large collection of untrimmed videos.
The former is known as single video moment retrieval (SVMR)~\cite{anne2017localizing,liu2021context,zheng2022progressive,qu2020fine,liu2020jointly,yang2022video,tang2024context,jiang2024sdn}, and the latter is known as video corpus moment retrieval (VCMR)~\cite{escorcia2019temporal,paul2021text,zhang2020hierarchical,wang2022siamese}.
In SVMR, existing methods mainly concentrate on how to precisely localize temporal boundings of target moments, and could be typically classified as proposal-based methods~\cite{chen2018temporally,yuan2019semantic,zhang2019man,wang2020temporally,gao2021fast,wang2021structured} and proposal-free methods~\cite{yuan2019find,wu2020tree,qu2020fine,fang2023you,zhang2023text,zhao2023diffusionvmr}.
Proposal-based methods first generate multiple moment proposals, then match them with a query to determine the most relevant one from the proposals. 
Without generating moment proposals, proposal-free methods predict the start and end time points of the target moment based on the fused video-query feature.
Although SVMR methods have achieved remarkable progress, there still exists a fatal defect that a corresponding video must be provided first to obtain the ground truth moment.
This setting fails to handle realistic scenarios where videos are not specifically provided and are generally collected in a large database.
Luckily, our PRVR task can serve as a reliable prior step by providing  videos that are partially related to a query from a large untrimmed video collection, thereby laying the groundwork for SVMR.

As the extension of SVMR, the VCMR task is to retrieve moments (or video segments) that are semantically relevant w.r.t. a given query from a collection of untrimmed videos \cite{escorcia2019temporal}. The typical methods for VCMR have a two-stage workflow. Firstly, multiple candidate videos that may contain the target moment are retrieved by global feature alignment between query and videos, followed by obtaining correct moments from the candidates utilizing moment selection approaches. 
Most VCMR methods focus on the latter localization stage. Early works\cite{escorcia2019temporal,paul2021text} adopt temporal convolutional module on candidate videos to gain proposals as abundant as possible.
For saving computation cost of generating proposals, Lei \etal\cite{lei2020tvr} constructed a novel Convolutional Start-End (ConvSE) detector which predicts the probability of each frame being the start or end of the target moment. Based on the model proposed by Lei \etal \cite{lei2020tvr}, Zhang \etal\cite{zhang2021video} apply contrastive learning on query-moment pairs, achieving higher performance. Yoon \etal \cite{yoon2022cascaded} built the proposal-free and proposal-based moment score maps concurrently to perform the more precise moment retrieval. Except for encoding moments and query separately, Zhang \etal\cite{zhang2020hierarchical} and Hou \etal \cite{hou2021conquer} introduce cross modal attention to modeling them in an interactive way, although they achieve relatively high performance, it is also time-consuming.

However, as mentioned above, most VMCR methods focus on obtaining the final relevant moments but ignore the importance of the intermediate step of finding more reliable candidate videos. Instead, our PRVR helps to retrieve more accurate partially related video from the video corpus in the first stage of VCMR.

\begin{figure*}[tb!]
\centering\includegraphics[width=2\columnwidth]{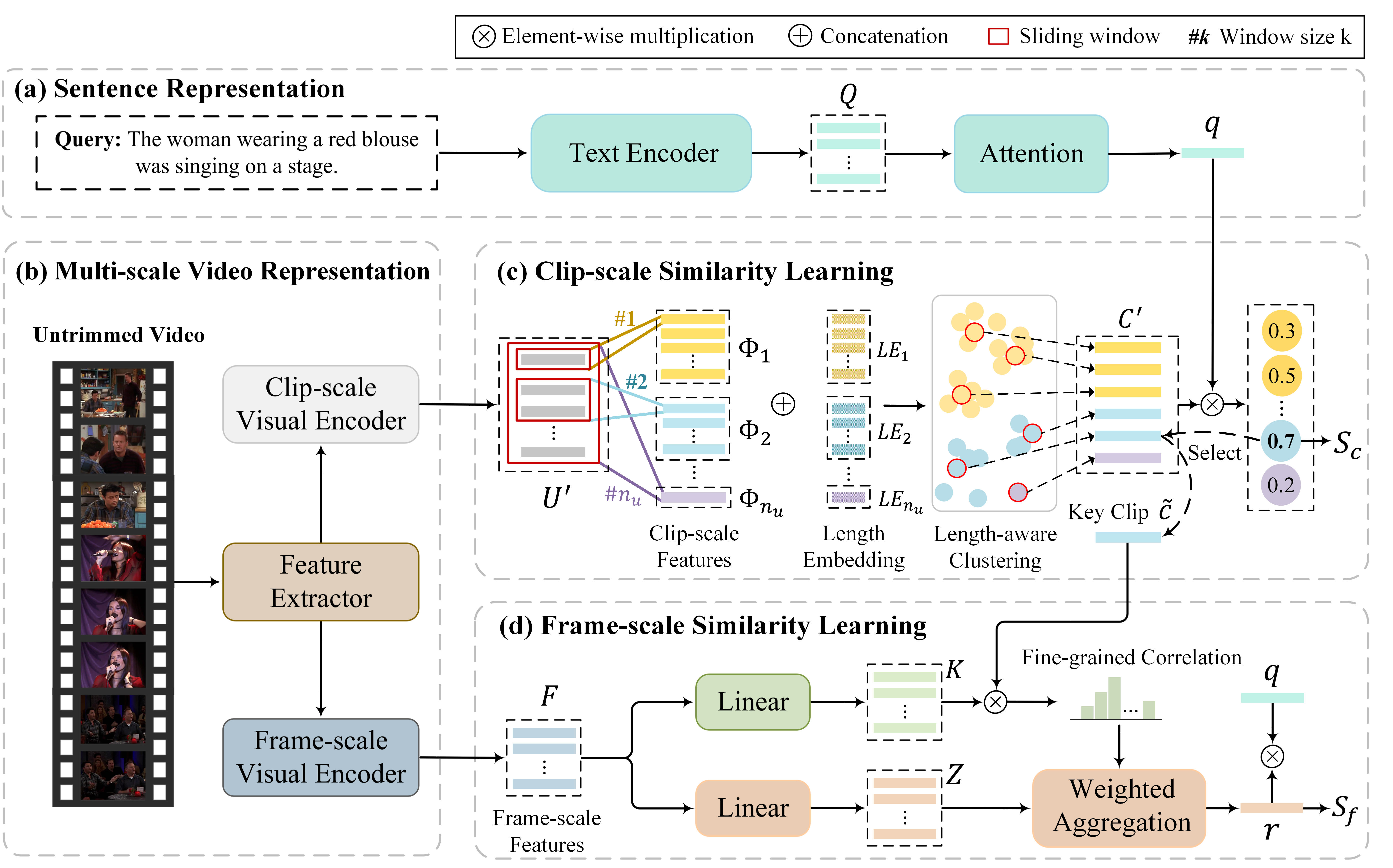}
\caption{The framework of our proposed model MS-SL++ for PRVR. Given an untrimmed video, the multi-scale video representation module encodes it into both clip-scale and frame-scale features. For more efficient coarse-to-fine similarity computation, a length-aware clustering module is applied to select representative clip-level features.
Meanwhile, the simple yet effective sentence representation module will encode the text query into a single feature vector. Then, the coarse relevance between the video and textual query is obtained by clip-scale similarity learning, whilst a key clip is detected. Moreover, the key clip is utilized for guiding the aggregation of frame-scale features and thus output the fine-grained relevance of the video and query. 
The partial similarity between an untrimmed video and a query is determined by the multi-scale similarities concurrently.
}\label{fig: model}
\end{figure*}

\section{Our Method}

\subsection{Overview}\label{sec:formulation}

\textbf{Definition of PRVR.}
Given a sentence-like query, 
\textit{PRVR} aims to retrieve target videos containing a specific moment  semantically relevant \wrt the given query, from a large corpus of untrimmed videos.
As the moment referred to by the query typically constitutes only a small portion of the target video with query-irrelevant content, we argue that the query is partially relevant to the video.
It is worth noting that PRVR is different from the current T2VR~\cite{dong2019dual,chen2020fine,han2021fine} task, where videos are pre-trimmed and much shorter, and queries are usually fully relevant to the whole video.
To build a PRVR model, a set of untrimmed videos is given for training, where each video is associated with multiple natural language sentences. Each sentence describes the content of a specific moment in the corresponding video. Note that we do not have access to the start and end time points of the target moments (moment annotations) referred to by the sentences.

\textbf{Overall pipeline.}
We propose a MS-SL++ network for the PRVR task. In particular, given an untrimmed video, we have no prior knowledge about where the relevant content is located, making it challenging to directly compute video-text similarity on a fine-grained scale.
Therefore, we propose formulating PRVR as a MIL problem, where a video is viewed as a bag of video clips and a bag of video frames simultaneously. 
We only need to find whether it has a text-related frame/clip to determine if a video is partially related to the text. Our whole pipeline is illustrated in Fig. \ref{fig: model}. 
Specifically, we first encode the input textual query into a global sentence representation, and encode the video into coarse clip-scale and fine-grained frame-scale representations. Then, we integrate the sentence representation with clip-scale video representation to determine whether there exists a partially relevant clip. To accelerate the multi-clip inference, we introduce a clip semantic clustering strategy to choose the most representative ones for clip-text matching. In addition to the computed clip-scale similarity, we further calculate the frame-scale partial relevance reasoning with the guidance of the key clips. At last, both clip-scale and frame-scale contexts contribute to the final retrieval result.

\subsection{Sentence Representation} \label{ssec:sent-rep}

As our focus is on the video representation and partially relevant learning, we adopt the sentence representation approach used in~\cite{lei2020tvr}, considering its good performance on the VCMR task.
Specifically, given a textual query
consisting of $n_q$ words, a pre-trained RoBERTa \cite{liu2019roberta} is first employed to extract word features.
Note that although the RoBERTa features have been contextualized, for better task-specific adaptation (\ie text-video semantic alignment), the features are further enhanced by a Transformer-based block as follows. A 
fully connected (FC) layer with a ReLU activation is utilized to map the word features into a lower-dimensional space.
The reduced features, together with
learned positional embeddings, are fed into 
a standard Transformer layer to obtain a new sequence of $d$-dimensional word feature vectors $Q=\{q_i\}_{i=1}^{n_q}\in\mathbb{R}^{{d}\times{n_q}}$. 
Within the Transformer, these features are fed to a multi-head attention layer followed by a feed-forward layer, with both layers connected via residual connection \cite{he2016deep} and layer normalization \cite{ba2016layer}.
Finally, a sentence-level representation $q\in\mathbb{R}^{d}$ is derived by employing a simple attention on $Q$:
\begin{equation}\label{eq:att}
\begin{aligned}
    q=\sum_{i=1}^{n_q} {\alpha_{i}^{q}}\times{q_{i}}, \quad
    \alpha^q=Softmax(w^\mathsf{T}  {Q} ),
\end{aligned}
\end{equation}
where $Softmax$ denotes softmax layer, $w\in\mathbb{R}^{d\times{1}}$ is trainable vector, and $\alpha^q \in\mathbb{R}^{{1 \times n_q} } $ indicates the attention vector.

\subsection{Multi-Scale Video Representation}\label{sec:video_rep}

Given an untrimmed video, we first represent it by a sequence of $d_v$-dimensional feature vectors $V\in\mathbb{R}^{d_v\times{n_v}}$, where $n_v$ denotes the number of the vectors.
The feature sequence is obtained by extracting frame-level features using a pre-trained 2D CNN, or extracting segment-level features using a pre-trained 3D CNN.
For the sake of simplicity, we regard $V$ as a sequence of frame-level features in the following description.
Based on $V$, we construct multi-scale video representation, jointly using a clip-scale similarity learning branch and a frame-scale similarity learning branch.

\subsubsection{Clip-scale video representation\label{sec_clip_scale}}
Before constructing video clips, we first downsample the input in the temporal domain to reduce the length of the feature sequence, which helps decrease the computational complexity of the model.
Specifically, given a sequence of frame feature vectors $V$ as input,
we downsample them into fixed $n_u$ feature vectors, where each feature vector is obtained by mean pooling over the corresponding multiple consecutive frame features.
Then, the video is represented by a sequence of new feature vectors $U\in\mathbb{R}^{d_v\times{n_u}}$. 
In order to make the features more compact, we employ an FC layer with a ReLU activation.
Moreover, in order to model temporal dependencies among the features in $U$, we use a standard Transformer with a learned positional embedding. Our preliminary experiment on the TVR dataset shows that the use of the Transformer is beneficial, improving R@1 from 12.8 to 13.6.
Formally, through an FC layer and a one-layer Transformer, we obtain $U'=\{u_i\}_{i=1}^{n_u}\in\mathbb{R}^{d\times{n_u}}$ by:
\begin{equation}
\begin{split}
    U'=Transformer(FC(U)+PE),
\end{split}
\end{equation}
where $PE$ denotes the output of the positional embedding.

For holistic clip construction, we employ flexible sliding windows on $U'$ along its temporal dimension with a stride of 1, as illustrated in Fig. \ref{fig: model} (c).
Given a sliding window of size $k$, a clip feature is obtained by mean pooling over the features within the window. The resulting feature sequence is denoted as $\Phi_k$ which consists of $n_{u}$\textit{-k+1} clips.
Consequently, by jointly employing sliding windows of varied sizes as $\{1, 2, ..., n_u\}$, we obtain a set of sequences $\{\Phi_1, \Phi_2, ..., \Phi_{n_u}\}$. 
Putting them together, we get the clip-scale video features
$C\in\mathbb{R}^{d\times{n_c}}$ by:
\begin{equation}\label{eq:sliding window}
\begin{split}
    C &= \{\Phi_1, \Phi_2, ..., \Phi_{n_u}\} \\ 
    & = \{\{c_1,...,c_{n_u}\},\{c_{n_u+1},...c_{2n_u-1}\},...,\{c_{n_c}\}\} \\
    & = \{c_1, c_2, ..., c_{n_c}\},
\end{split}
\end{equation}
where $c_i \in\mathbb{R}^d$ denotes the feature representation of the $i$-th clip, $n_c$ is the number of all clip-scale video features which meets ${n_c}={n_u}({{n_u}+1})/2$.

\subsubsection{Frame-scale video representation}
Opposite to the clip-scale feature, the frame-scale feature is regarded as fine-grained temporal granularity, so we keep the initial features to preserve the detailed content of the video. However, as the initial frame features are extracted independently, they naturally lack temporal dependencies. To reintroduce these dependencies, we again employ Transformer to learn the self-contexts.
Specifically, given the frame feature sequence $V$, we first utilize an FC layer with ReLU activation to reduce the dimensionality of the input, followed by a standard Transformer with a positional embedding layer. The re-encoded frame features, denoted $F=\{f_i\}_{i=1}^{n_v}\in\mathbb{R}^{d\times{n_v}}$, are computed as: 
\begin{equation}
        F = Transformer(FC(V)+PE).
\end{equation}

Note that the network structures of Transformer, FC and PE are the same as those in the clip-scale visual encoder, but their trainable parameters are not shared. This allows each visual encoder to learn parameters suitable for their own scale.

\subsection{Multi-Scale Similarity Learning}\label{ssec:mssl}
According to the definition that a video is considered partially related to the text if it contains a text-related frame/clip, it is necessary to match the video-query pair at fine-grained clip and frame scales rather than using coarse global matching. As we have no prior knowledge about the localization of relevant content in PRVR, directly computing video-text similarity on a fine-grained scale is challenging. To this end, we propose a multi-scale similarity learning approach, which computes the similarity at different scales in a coarse-to-fine manner.
Specifically, it first detects a key clip that is most likely to be relevant to the query. 
Then, the importance of each frame is measured on a fine-grained temporal scale under the guidance of the key clip.
The final similarity is computed by jointly considering the query's similarities with both the key clip and the frames.
The framework of the multi-scale similarity learning is illustrated in Fig. \ref{fig: model} (c) and (d).

\subsubsection{Clip-scale similarity learning}\label{ssec:clip}
Recall that the clip-scale video features in $C$ are generated by sliding windows of varied sizes, resulting in clips of varying lengths, high temporal overlap, and redundancy in information. For each video, all clip-scale features have to be stored for inference, resulting in heavy memory consumption for large-scale video retrieval. All these issues hinder the deployment of retrieval models in real-world applications.

For faster inference,
we propose selecting from $C$ a subset of representative features $C'$ 
as the clip-scale video representation. To obtain $C'$, a straightforward idea is to conduct a clustering algorithm over $C$. 
However, as the clip-
scale features correspond to video clips of varying lengths,
such vanilla clustering strategy that does not account for
clip lengths is suboptimal. 
To improve the diversity of the lengths of the video clips associated with the representative features, we introduce a length-aware clustering strategy, which incorporates a length embedding to enhance the length information of $C$ when clustering.

Concretely, we firstly construct a length embedding matrix $LE\in\mathbb{R}^{{d_e}\times{l}}$ by the sine and cosine functions, like the position embedding used in Transformer~\cite{vaswani2017attention}, where $d_e$ is the embedding dimension and $l$ is the maximum length of clips.
Given the original clip representation $C$, the length-enhanced clip representation is obtained as
\begin{equation}
   \quad C^e=\{c_{1}^{e},c_{2}^{e},...,c_{n_c}^{e}\}, c_{i}^{e}=[c_i,LE_t], 
\end{equation}
where $LE_t\in\mathbb{R}^{d_e}$ is length embedding vector of length $t$ selected from $LE$, $[,]$ denotes the concatenation operation. Then a general clustering algorithm k-medoids algorithm~\cite{zhao2022centerclip} is conducted over the length-enhanced clip-scale representation $C^e$. This algorithm partitions data into $k$ clusters by minimizing the sum of distances between all data points in each cluster and the center of that cluster.
The center of a specific cluster is defined as an actual data point that has a minimal average distance to all other points within the same cluster.
Subsequently, the center feature of clusters is kept as the representative feature, while all other non-center features are dropped to avoid duplicated computation of similar clip-scale features. After the clustering, we get $C'=\{c'_i\}_{i=1}^{n_k}\in\mathbb{R}^{{d}\times{n_k}}$ key clips, where $n_k$ denotes the number of clusters.

Then we can construct query-relevant similarity learning on $C'$. We compute cosine similarities between
$C'$ and the query representation $q$,
followed by a max-pooling operator on the similarities. 
More formally, the clip-scale similarity is obtained as:
\begin{equation}
\begin{aligned}
     S_c(v, q) = \mathop{\max}(\{cos(c'_1, q), cos(c'_2, q),...,cos(c'_{n_k}, q)\}),\\
\end{aligned}
\end{equation}
where $cos(\cdot)$ denotes the cosine similarity function.
The max-pooling operation could select the clip-scale feature that owns the highest semantic relevance with the query, so we treat it as \emph{key clip} feature for the follow-up frame-scale similarity learning.
It is worth noting that although the max-pooling operation is simple, it allows the model to identify partial relevance while ignoring irrelevant segments.

\subsubsection{Frame-scale similarity learning}\label{sec:frame-scale similarity}
We hypothesize that if the model has a preliminary understanding of the  coarse relevant content with respect to the query, it will enhance the model's ability to accurately locate more relevant content at a finer scale. Thus we devise a Key Clip Guided Attention (KCGA) mechanism that leverages the idea of multi-head self-attention (MHSA) mechanism in Transformer \cite{vaswani2017attention} for frame-scale similarity learning. It can explore frames that are semantically close to the key clip obtained in Section \ref{ssec:clip} and supplement detailed information of the key clip in the video. Specifically, given a sequence of frames features $F = \{f_1, f_2, ..., f_{n_v}\}$ as input, the original
MHSA first projects them into queries, keys, values, and then computes the output as a weighted sum of the values. The weight assigned to each value is computed by a compatibility function of the query with the corresponding key.
Different from MHSA utilizing the same input to construct queries, keys, and values here we take the feature vector of the key clip as the query, and the video frame features as keys and values. Formally, the aggregated frame feature vector is obtained as:
\begin{equation}
\begin{aligned}
     r = Softmax(\tilde{c}^\mathsf{T}K)Z^\mathsf{T}, \quad
     K = W_k{F}, \quad
     Z = W_z{F}, 
\end{aligned}
\end{equation}
where $\tilde{c}\in\mathbb{R}^{d\times{1}}$ indicates the feature vector of the key clip, $W_k\in\mathbb{R}^{d\times{d}}$ and $W_z\in\mathbb{R}^{d\times{d}}$ are two trainable projection matrices.
The $\tilde{c}^\mathsf{T}K$ measures the fine-grained correlation between frames and the key clip, which is expected to yield larger values for frames that are more similar to the key clip and suppress the influence of irrelevant background frames. Therefore, frames that are more similar to the key clip will receive higher attention weights.

Finally, the frame-scale similarity is measured as the cosine similarity between the aggregated frame feature vector $r$ and the query feature vector $q$, namely:
\begin{equation}
\begin{aligned}
     S_f(v, q) = cos(r, q).\\
\end{aligned}
\end{equation}

\subsection{Model Training and Inference}\label{sec:model_infer}
\subsubsection{Training stage}
In this section, we first introduce the definition of positive and negative pairs for model training.
Inspired by MIL \cite{dietterich1997solving,maron1997framework}, we define that a query and video pair is positive if the video contains certain content that is relevant to the query, and negative if no relevant content in the video. 

Based on the above definition, we jointly use the triplet ranking loss \cite{faghri2017vse++,dong2021dual} and infoNCE loss \cite{miech2020end,zhang2021video} that are widely used in retrieval related tasks, and find them complementary. 
Given a positive video-query pair $(v, q)$, the triplet ranking loss over the mini-batch $\mathcal{B}$ is defined as:
\begin{equation}
\begin{split}
    \mathcal{L}^{trip}=\frac{1}{n}\sum_{(q,v) \in \mathcal{B}}[max(0,m+{S}(q^-, v)-{S}(q, v))\\
    +max(0,m+{S}(q, v^-)-{S}( q,v))],
\end{split}
\end{equation}
where $m$ is the margin constant, $S(\cdot)$ denotes the similarity function which we can use the clip-scale similarity $S(\cdot)_c$ or the frame-scale similarity $S(\cdot)_f$. Besides, $q^-$ and $v^-$ respectively indicate a negative sentence sample for $v$ and a negative video sample for $q$. The negative samples are randomly sampled from the mini-batch at the beginning of the training, while being the hardest negative samples after 20 epochs.

Given a positive video-query pair $(v, q)$, the infoNCE loss over the mini-batch $\mathcal{B}$ is computed as:
\begin{equation}
\begin{aligned}
    \mathcal{L}^{nce}=-\frac{1}{n}\sum_{(q,v) \in \mathcal{B}} \left[ 
     {log\left({\frac{{S}(q,v)}{{S}(q,v)+\sum_{q^-_i\in\mathcal{N}_q}{S}(q^-_i, v)}}\right)} \right. \\
    \left. +{log\left({\frac{{S}(q,v)}{{S}(q,v)+\sum_{v^-_i\in\mathcal{N}_v}{S}(q, v^-_i)}}\right)} 
    \right],
\end{aligned}
\end{equation}
where $\mathcal{N}_q$ denotes all negative queries of the video $v$ in the mini-batch, while $\mathcal{N}_v$ denotes all negative videos of the query $q$ in the mini-batch.

As the previous work~\cite{li2020sea} has concluded that using one loss per similarity function performs better than using one loss on the summation of multiple similarities, we employ the above two losses on both clip-scale similarity and frame-scale similarity, instead of their sum.
Finally, our model is trained by minimizing the following overall training loss:
\begin{equation}
    \mathcal{L}=\mathcal{L}^{trip}_c+ \mathcal{L}^{trip}_f+\lambda_{1}\mathcal{L}^{nce}_{c}+\lambda_{2}\mathcal{L}^{nce}_{f},
\end{equation}
where $\mathcal{L}^{trip}_c$ and $\mathcal{L}^{trip}_f$ denote the triplet ranking loss using the clip-scale similarity $S(\cdot)_c$ and frame-scale similarity $S(\cdot)_f$ respectively, and accordingly for $\mathcal{L}^{nce}_{c}$ and $\mathcal{L}^{nce}_{f}$. $\lambda_{1}$ and $\lambda_{2}$ are hyper-parameters to balance multiple losses.

\subsubsection{Inference stage}
After the model has been trained, the partial similarity between an untrimmed video and a textual query is computed as the sum of their clip-scale similarity and frame-scale similarity, namely:
\begin{equation}\label{eq:alpha}
    S(v,s)= \alpha S_c(v,s) + (1-\alpha) S_f(v,s)
\end{equation}
where $\alpha$ is a hyper-parameter to balance the importance of two similarities, ranging within [0, 1].
Given a query, we sort all videos from the video gallery in descending order according to their similarity with respect to the given query.

\section{Experiments} \label{sec:eval}

\subsection{Experimental Setup} \label{ssec:exp-set}

\subsubsection{Datasets}
In order to verify the viability of our proposed model for PRVR, queries that are partially relevant to videos are required. As videos in popular T2VR datasets such as MSR-VTT \cite{xu2016msr}, MSVD \cite{chen2011collecting} and VATEX \cite{wang2019vatex} are supposed to be fully relevant to the queries, they are not suited for our experiments.
Here, we re-purpose three datasets commonly used for VCMR, \ie TVR \cite{lei2020tvr}, ActivityNet-Captions \cite{krishna2017dense}, and Charades-STA \cite{gao2017tall},
considering their natural language queries partially relevant to the corresponding videos (a query is typically associated with a specific moment in a video).
Table \ref{tab:dataset_stat} summarizes the brief statistics of these datasets, including average lengths of moments and videos, and the average moment length proportion in the whole video (moment-to-video ratio). Note that as we focus on retrieving videos, moment annotations provided by these datasets are not used in our newly proposed PRVR task.

\begin{table} [tb!]
\renewcommand{\arraystretch}{1.2}
\caption{Brief statistics of three public datasets used in our experiments.
The length is measured in seconds.}
\label{tab:dataset_stat}
\centering 
\scalebox{0.85}{
\begin{tabular}{lrrrrrrr}
\toprule
\multirow{2}{*}{\textbf{Datasets}} & \multicolumn{1}{c}{} & \multicolumn{2}{c}{\textbf{Average length}} & \multicolumn{1}{l}{} & \multicolumn{3}{l}{\textbf{Moment-to-video ratio}} \\ \cline{3-4} \cline{6-8} 
 &  & moments & videos & \multicolumn{1}{l}{} & \multicolumn{1}{r}{\emph{min}} & \multicolumn{1}{r}{\emph{max}} & \multicolumn{1}{r}{\emph{mean}} \\ \hline
TVshow Retrieval &    & 9.1  & 76.2  & & 0.48\% & 100\%  & 11.9\% \\
\multicolumn{1}{c}{ActivityNet-Captions} &   & 36.2  & 117.6 &  & 0.48\% & 100\% & 30.8\% \\
Charades-STA &   & 8.1   & 30.0 & & 4.3\% & 100\%  & 26.3\% \\
\bottomrule
\end{tabular}
 }
\end{table}

\textbf{TVshow Retrieval (TVR)} \cite{lei2020tvr} is a multimodal dataset originally for video corpus moment retrieval, where videos are paired with subtitles generated by automatic speech recognition.
It contains 21.8K videos collected from 6 TV shows. Each video is associated with 5 sentences that describe a specific moment in the video. As such a sentence is partially relevant to the video.
Following \cite{zhang2020hierarchical,zhang2021video}, we utilize 17,435 videos with 87,175 moments for training and 2,179 videos with 10,895 moments for testing. 

\textbf{ActivityNet-Captions} \cite{krishna2017dense} is originally developed for dense video captioning, and is now a popular dataset for SVMR.
It contains around 20K videos from Youtube, and the average length of videos is the largest among the datasets we used. On average, each video has around 3.7 moments with corresponding sentence descriptions. We use the popular data partition used in \cite{zhang2020hierarchical, zhang2021video}.

 \textbf{Charades-STA} \cite{gao2017tall} contains 6,670 videos with 16,128 sentence descriptions. Each video has around 2.4 moments with corresponding sentence descriptions on average. We utilize the official data partition for model training and evaluation.

\subsubsection{Evaluation metrics}
To evaluate PRVR models, we utilize rank-based metrics, namely $R@K$ ($K = 1, 5, 10, 100$) and Median rank (Med r), which are commonly used for current T2VR task~\cite{wang2020learning,dong2021dual}. 
$R@K$ is the fraction of queries that correctly retrieve desired items in the top $K$ of the ranking list. 
The performance is reported in percentage (\%). Med r is the median rank of the first relevant item in the search results.
Higher $R@K$ and lower Med r means better performance. For overall comparison, we also report the Sum of all Recalls (SumR).

\subsubsection{Implementation details}
Following previous works \cite{lei2020tvr,zhang2021video}, we utilize CNN-based features to represent videos.
On TVR, we utilize the feature provided by \cite{lei2020tvr}, that is, 3,072-D  visual features obtained by the concatenation of frame-level ResNet152 \cite{he2016deep} feature and segment-level I3D \cite{carreira2017quo} feature.
For ease of reference, we refer to it as ResNet152-I3D.
On ActivityNet-Captions and  Charades-STA, we utilize the I3D features, which are respectively provided by \cite{zhang2020hierarchical} and \cite{mun2020local}.
For the sentence feature, we use the 768-D RoBERTa feature provided by \cite{lei2020tvr} on TVR, where RoBERTa is fine-tuned on the queries and subtitle sentences of TVR.
On ActivityNet-Captions and Charades-STA, we use the 1,024-D RoBERTa feature extracted by ourselves using the open RoBERTa toolkit\footnote{\url{https://pytorch.org/hub/pytorch_fairseq_roberta/}}.
Additionally, we observe an increasing use of Contrastive Language-Image Pre-Training (CLIP)~\cite{radford2021learning,gorti2022x} for visual and text representation, we also use it in our experiments. We utilize the visual branch and the textual branch of CLIP (ViT-B/32) to extract frame and sentence features respectively, and freeze them without fine-tuning.

For the video representation module, we set the fixed number $n_u$ to 32 in the downsampling strategy.
Besides, we set the maximum frame number $n_v$ to 128. Once the number of frames is over $n_v$, it will be downsampled to $n_v$.
For sentences, we set the maximum length of query $n_q$ to 32 on TVR and Charades-STA, 64 on ActivityNet-Captions, and the words outside the maximum length are simply discarded.
For the Transformer module used in our model, we set its hidden size $d=384$, and 4 attention heads are employed.
For the k-medoids algorithm, we utilize Euclidean distance to measure distances between data points.

For hyper-parameters in the loss functions, we empirically set $\lambda_{1}$=0.03 and $\lambda_{2}$=0.04 to make all loss elements have a similar loss value at the beginning of the model training.
For model training, we utilize an Adam optimizer with a mini-batch size of 128. 
The initial learning rate is set to 0.00025, and we take a learning rate adjustment schedule similar to \cite{lei2020tvr}. Early stop occurs if the validation performance does not improve in ten consecutive epochs. The maximal number of epochs is set to 100.

\subsection{Ablation Studies} \label{ssec:ablation}

In this section, we evaluate the effectiveness of each component of our
proposed model on the TVR dataset.

\begin{table} [tb!]
\renewcommand{\arraystretch}{1.2}
\caption{Ablation study on the TVR dataset.}
\label{tab:ablation}
\centering 
\scalebox{0.82}{
\begin{tabular}{lcccccc}
\toprule
\textbf{Model} & \textbf{R@1} & \textbf{R@5} & \textbf{R@10} & \textbf{R100} &  \textbf{SumR}$\uparrow$ &\textbf{Med r $\downarrow$}\\
\hline
Full setup & \textbf{13.6} & \textbf{33.1} & \textbf{44.2} & \textbf{83.5} & \textbf{174.4} & \textbf{14}\\
\hline
w/o frame-scale branch & 12.5 & 31.4 & 41.8 & 82.2 & 167.9  & 18\\
w/o clip-scale branch & 8.0 & 21.0 & 30.0 & 74.0 & 133.0 &33\\
\hline 
w/o key clip guide & 12.3 &	30.8 &	41.4 &	82.0 &	166.5 & 17\\
w/o weighted aggregation & 12.0 &	30.2 &	40.8 &	81.5 &	164.5 & 17\\
\hline
w/o infoNCE &11.5  & 29.4 & 40.5 & 81.2 & 162.6  & 19\\
w/o Triplet loss & 11.4 & 29.8 & 40.9 & 81.9 & 163.9 & 19\\
\bottomrule
\end{tabular}
}
\end{table}

\subsubsection{The effectiveness of multi-scale similarity learning branches}
To examine the usefulness of the multi-scale branches, we compare the counterparts without the clip-scale similarity learning branch or the frame-scale similarity learning branch.
As shown in Table \ref{tab:ablation}, removing any branch results in clear performance degeneration. The result not only demonstrates the effectiveness of the multi-scale solution, but also shows the complementarity between the clip-scale and the frame-scale similarity learning branches.

\begin{table}[tb!]
\renewcommand{\arraystretch}{1.2}
\caption{The influence of the cluster number $n_k$ on model performance and inference time.
The last line is the result of our model without using the clustering for acceleration.
}\label{tab:cluster_number}
\centering
\begin{tabular}{llccccc}
\toprule
\textbf{$n_k$} & \textbf{} & 
\textbf{\begin{tabular}[c]{@{}c@{}}Performance\\ (SumR) $\uparrow$ \end{tabular}} & \textbf{} & \textbf{\begin{tabular}[c]{@{}c@{}}Retrieval Time\\ (ms) $\downarrow$ \end{tabular}} & \textbf{} & \textbf{\begin{tabular}[c]{@{}c@{}}Feature Storage \\ (number) $\downarrow$ \end{tabular}}
\\ 
\hline
8 &  & 167.3 &  & 10 &  & 59 \\
16 &  & 171.1 &  & 11 &  & 67 \\
32 &  & 174.5 &  & 12 &  & 83 \\
64 &  & 173.2 &  & 16 &  & 115 \\
128 &  & 172.4 &  & 21 &  &  179 \\
256 &  & 173.7 &  & 28 &  &  307 \\ \hline
- &  & 172.4 &  & 33 &  & 579 \\ 
\bottomrule
\end{tabular}
\end{table}

\subsubsection{The effectiveness of key clip guided attention}
Recall that our proposed key clip guided attention is mainly comprised of the key clip guide and the weighted aggregation.
To evaluate their contributions, we first compare the variant w/o key clip guide, which is implemented by replacing the key clip guided attention with simple attention. The simple attention is implemented as Eq. \ref{eq:att} without any guidance.  Additionally, we further investigate a variant w/o weighted aggregation, where the weighted aggregation is substituted with a mean-pooling strategy.
As shown in Table \ref{tab:ablation}, our model with the full setup outperforms both ablated variants (R@1 of 13.6 vs. 12.3 and 12.0, respectively). 
These results not only confirm the effectiveness of our key clip guided attention design but also highlight the importance of both the key clip guide and the weighted aggregation.

\subsubsection{The effectiveness of the combination of triplet ranking loss and infoNCE loss}
To validate the choice of joint use of the two losses, we compare the results of using either triplet ranking loss or infoNCE loss. As shown in Table \ref{tab:ablation}, triplet ranking loss and infoNCE give comparable results when used alone, but they are much worse than the model with the full setup of jointly using both. 
The result demonstrates the benefit of using these two losses jointly.

\begin{figure}[tb!]
\subfigure[]{
\includegraphics[width=0.47\columnwidth]{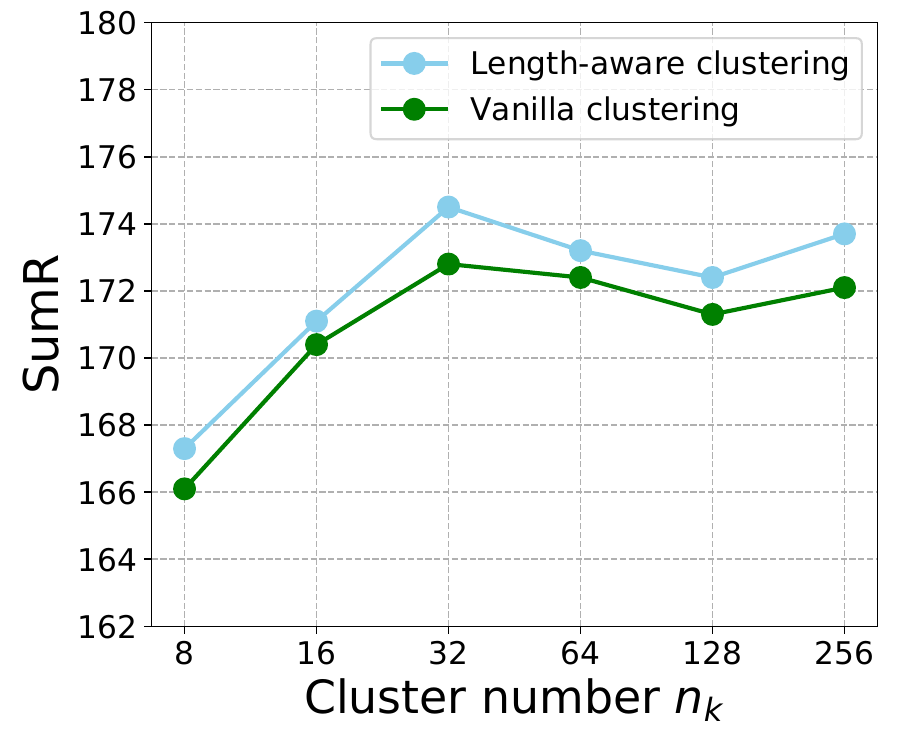}
}
\subfigure[]{
\includegraphics[width=0.47\columnwidth]{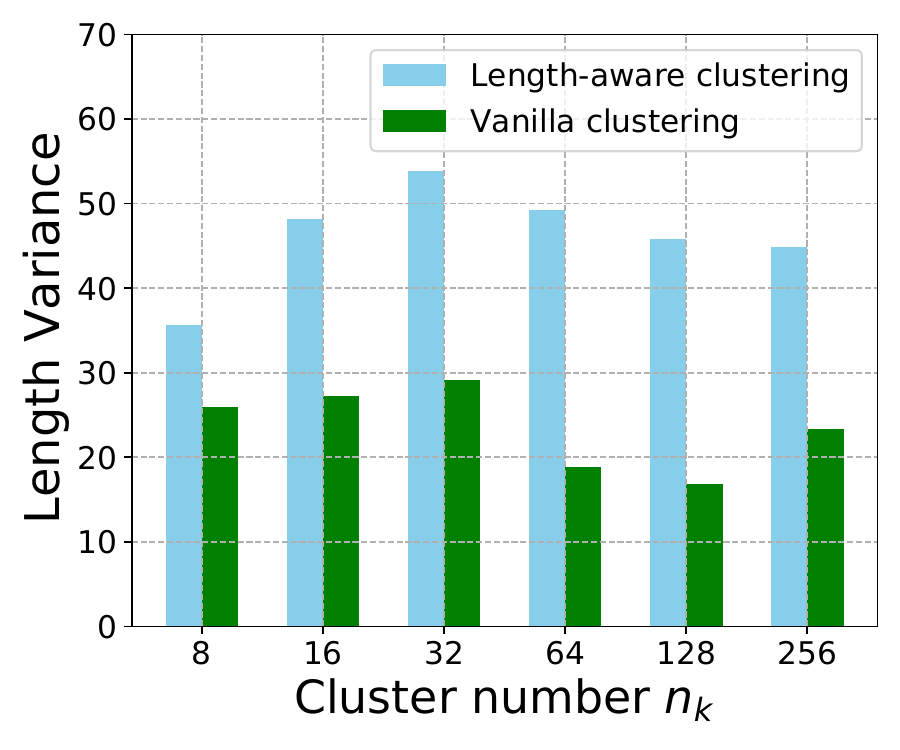}
}
\caption{Comparison between length-aware clustering and vanilla clustering in terms of their (a) retrieval performance and (b) the length diversity of the obtained representative clips by clustering.
}\label{fig:effect-length-emb}
\end{figure}

\subsubsection{The studies on length-aware clustering}\label{sec:n_k}
\textbf{The effect of cluster's number} $\boldsymbol{n_k}$.
We first study the effect of cluster’s number (\ie the number of representative video clips) on the model performance, retrieval time and feature storage.
For retrieval time, we omit the time required for video representation as videos can be represented offline in advance. The time is evaluated as the average time of answering 10,895 queries on the TVR dataset.
Regarding feature storage, we measure it as the number of clip-level video features extracted offline for each video.
The results with respect to the number of clusters are summarized in Table \ref{tab:cluster_number}.
When the number of clusters $n_k$ decreases, both retrieval time and feature storage are shortened. However, we observe that a small number of clusters, $e.g.$, 8 or 16, results in obvious performance degradation. We attribute it to the fact that a small number of clusters may lead to the representative clips being insufficient thus failing to adequately represent the content of the original video. When the number of clusters is larger than 16, the performance is comparable to the model without using clustering for acceleration. The results demonstrate the effectiveness of our proposed clustering strategy for acceleration.
Note that using the number of clusters as 32 achieves the best balance of model performance and retrieval cost, so we use it as the default value unless otherwise stated.

\textbf{The effectiveness of length-aware clustering}.
To verify the effectiveness of length-aware clustering in the selection of representative clips, we conduct a comparison with vanilla clustering that does not incorporate length embedding.
As shown in Fig. \ref{fig:effect-length-emb} (a), given varied cluster numbers, length-aware clustering consistently outperforms vanilla clustering, demonstrating its effectiveness for representative clip selection.
The better performance can be attributed to the fact that length-aware clustering takes clip length into account, which allows the representative clips obtained by clustering to be more diverse in terms of their length.
To further confirm the diversity of the representative clips obtained by length-aware clustering, we calculate the clip length variance of all videos from the training set of the TVR dataset. The results, depicted in Figure \ref{fig:effect-length-emb}, clearly show that length-aware clustering leads to a higher length variance compared to vanilla clustering.

\subsubsection{Clip granularity sensitivity analysis}\label{sec:clip-rep}

\begin{figure}[tbp!]
\centering\includegraphics[width=0.6\columnwidth]{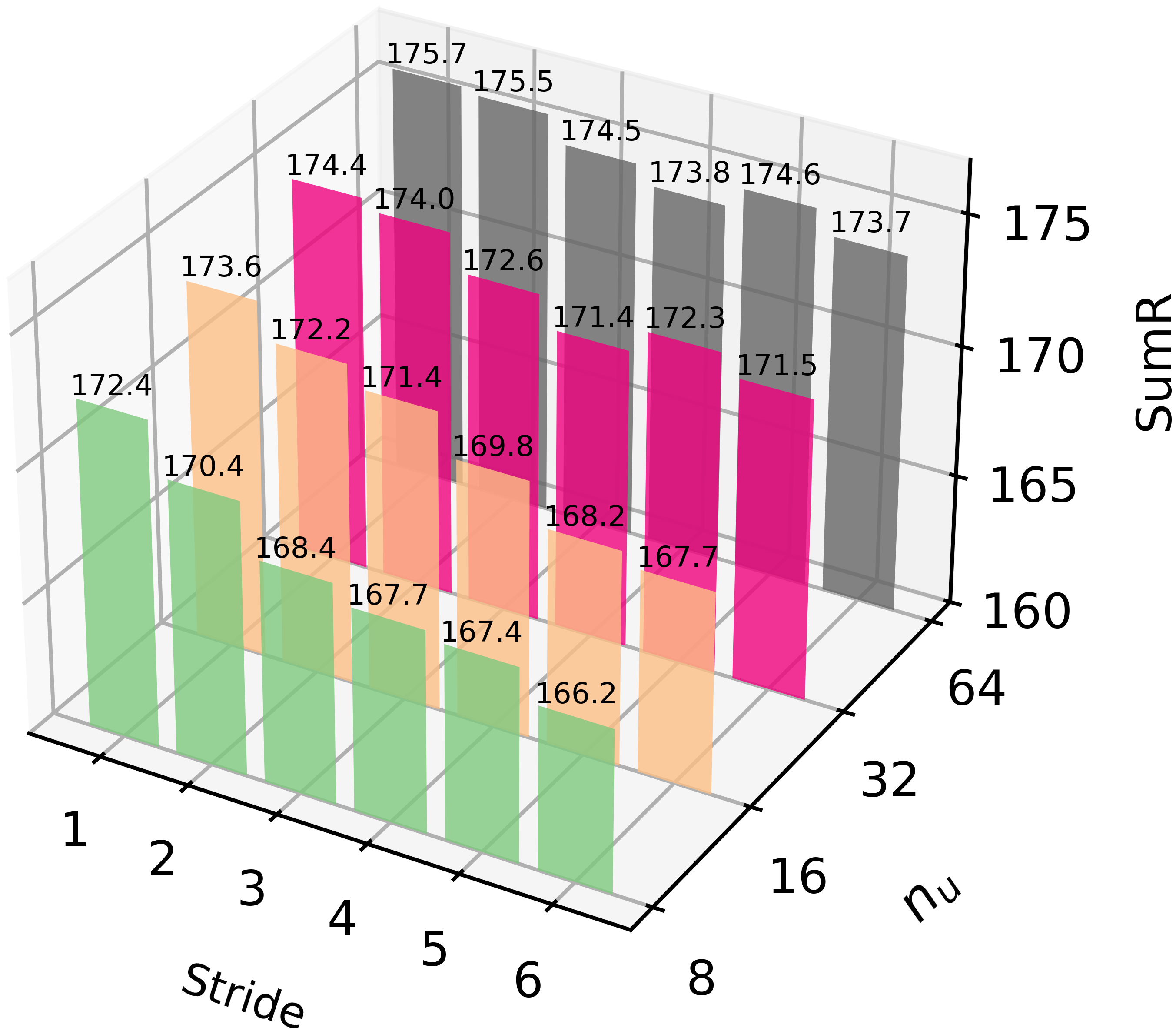}
\caption{Impact of clip granularity on retrieval performance. 
}\label{fig:stride}
\end{figure}

We check how different granularity levels of the generated clips impact 
performance. Recall that the clip granularity is jointly determined by the number of sliding windows ($n_u$) and their stride. As Fig. \ref{fig:stride} shows, more fine-grained clips (with larger $n_u$ and smaller stride) mean better performance, yet at the cost of higher computational overhead. 
As a trade-off between effectiveness and efficiency, our default choices of $n_u$ and stride are 32 and 1, respectively.

\begin{table}[tbp!]
\renewcommand{\arraystretch}{1.3}
\centering
\caption{
Comparison between our proposed MS-SL++ with simple baselines that rely only on CLIP features without additional model components.
}
\scalebox{0.72}{
\begin{tabular}{llcccccc}
\toprule
\textbf{Dataset} & \multicolumn{1}{l}{\textbf{Method}} & \multicolumn{1}{l}{\textbf{R@1}} & \multicolumn{1}{l}{\textbf{R@5}} & \multicolumn{1}{l}{\textbf{R@10}} & \multicolumn{1}{l}{\textbf{R@100}} & \multicolumn{1}{l}{\textbf{SumR}} & \multicolumn{1}{l}{\textbf{Med r}} \\ \hline
\multirow{4}{*}{TVR}  &CLIP-\textit{cluster}  & 9.0 & 19.8 & 26.7 & 60.5 & 116.0 & 54 \\
 & CLIP-\textit{psd}  & 6.8 & 16.6 & 22.7 & 56.6 & 102.7 & 72 \\
 & CLIP-\textit{svmr}  & 8.8 & 19.8 & 26.8 & 60.5 & 115.9 & 54 \\ 
 & MS-SL++  & \textbf{23.4} & \textbf{47.2} & \textbf{58.3} & \textbf{90.1} & \textbf{219.0} & \textbf{6} \\ \hline
\multirow{4}{*}{ActivityNet}& CLIP-\textit{cluster}  & 13.0 & 29.6 & 40.2 & 74.7 & 157.4 & 19 \\
 & CLIP-\textit{psd} & 11.8 & 27.7 & 38.4 & 73.1 & 150.9 & 21 \\
 & CLIP-\textit{svmr}  & \textbf{13.5} & 30.4 & 41.5 & 78.2 & 163.6 & 18 \\ 
 & MS-SL++  & 12.7 & \textbf{33.1} & \textbf{46.0} & \textbf{82.1} & \textbf{173.9} &\textbf{13} \\ \hline
 \multirow{4}{*}{Charades-STA}  & CLIP-\textit{cluster}  &0.9	&4.2	&7.0	&30.3	&42.4	&268 \\
 & CLIP-\textit{psd}  &1.0	&3.9	&6.7	&27.8	&39.5 &298 \\
 & CLIP-\textit{svmr}  &1.2	&4.7	&7.5	&32.0	&45.4 &258\\ 
 & MS-SL++  & \textbf{1.5} & \textbf{5.2} & \textbf{9.3} & \textbf{39.2} & \textbf{55.2} &\textbf{171} \\ 
\bottomrule
\end{tabular}
}\label{tab:zero-shot eval}
\end{table}

\subsection{Comparison with Simple Baselines}\label{sec:zs eval}
To further verify the proposed framework, we compare with a simple baseline that relies only on frame-level CLIP features without additional model components. 
Depending on how \emph{candidate moments} are generated for the given video, we implement the simple baseline in three manners: 
\begin{itemize}
    \item CLIP-\emph{cluster}, generating moments by our proposed length-aware clustering method. 
    \item CLIP-\emph{psd}, generating moments using PySceneDetect\footnote{\url{https://github.com/Breakthrough/PySceneDetect}}, a public shot boundary detection tool.
    \item CLIP-\emph{svmr}, generating moments using UniVTG~\cite{lin2023univtg}, a supervised SVMR model that has been pre-trained on multiple SVMR datasets for predicting query-dependent temporal boundaries. 
\end{itemize}
Once the temporal boundaries of the moments are determined, we obtain moment-level features by average pooling over the corresponding frame-level features. The similarity between each moment and the text query is then measured using cosine similarity, and the highest similarity score among all moments is taken as the final video-text relevance score for ranking.

As shown in Table \ref{tab:zero-shot eval}, we report the best-performing configuration for each method on the three benchmarks. Among the three baselines,  CLIP-\emph{svmr} shows relatively strong performance, especially on ActivityNet, likely due to its prior training on supervised grounding tasks. CLIP-\emph{cluster} using our proposed length-aware clustering consistently outperforms CLIP-\emph{psd} using PySceneDetect, demonstrating the effectiveness of our clip selection strategy. 
MS-SL++ achieves the best overall performance in terms of SumR and Med r. 
These results highlight the necessity of learning-based optimization, as well as the benefit of modeling fine-grained partial similarities for the PRVR task.

\begin{table*}[tb!]
\renewcommand{\arraystretch}{1.2}
\caption{Performance of PRVR on the three PRVR benchmarks. 
Models are sorted in ascending order in terms of their overall performance. \\
}
\label{tab:sota}
\centering 
\scalebox{0.73}{
\begin{tabular}{lccccccccccccccccccccc}
\toprule
\multirow{2}{*}{\textbf{Method}} &  & \multicolumn{6}{c}{\textbf{TVR}} &  & \multicolumn{6}{c}{\textbf{ActivityNet-Captions}} &  & \multicolumn{6}{c}{\textbf{Charades-STA}} \\ 
\cline{3-8} \cline{10-15} \cline{17-22} 
 &  & R@1 & R@5 & R@10 & R@100 & SumR & Med r & & R@1 & R@5 & R@10 & R@100 & SumR & Med r &  & R@1 & R@5 & R@10 & R@100 & SumR & Med r\\ \cline{1-1} \cline{3-8} \cline{10-15} \cline{17-22} 
\textbf{CNN-based features:} &  &  &  &  &  &  &  &  &  &  &  &  &  &  &  &  &  &  \\
W2VV \cite{dong2018predicting} &  & 2.6 & 5.6 & 7.5 & 20.6 & 36.3 &552 &  & 2.2 & 9.5 & 16.6 & 45.5 & 73.8 &134 &  & 0.5 & 2.9 & 4.7 & 24.5 & 32.6 &541 \\
HGR \cite{chen2020fine} &  & 1.7 & 4.9 & 8.3 & 35.2 & 50.1 &262 &  & 4.0 & 15.0 & 24.8 & 63.2 & 107.0 &41 & & 1.2 & 3.8 & 7.3 & 33.4 & 45.7 &263\\
HTM \cite{miech2019howto100m} &  & 3.8 & 12.0 & 19.1 & 63.2 & 98.2 &59 &  & 3.7 & 13.7 & 22.3 & 66.2 & 105.9 &44 &  & 1.2 & 5.4 & 9.2 & 44.2 & 60.0 &125 \\
CE \cite{liu2019use} &  & 3.7 & 12.8 & 20.1 & 64.5 & 101.1 &65 &  & 5.5 & 19.1 & 29.9 & 71.1 & 125.6 &26 &  & 1.3 & 4.5 & 7.3 & 36.0 & 49.1 &210\\
W2VV++ \cite{li2019w2vv++} &  & 5.0 & 14.7 & 21.7 & 61.8 & 103.2 &62 &  & 5.4 & 18.7 & 29.7 & 68.6 & 122.6 &30 &  & 0.9 & 3.5 & 6.6 & 34.3 & 45.3 &265\\
VSE++ \cite{faghri2017vse++} &  & 7.5 & 19.9 & 27.7 & 66.0 & 121.1 &44 &  & 4.9 & 17.7 & 28.2 & 66.4 & 117.8 &32 &  & 0.8 & 3.9 & 7.2 & 31.7 & 43.6 &281\\
DE \cite{dong2019dual} &  & 7.6 & 20.1 & 28.1 & 67.5 & 123.4 &43 &  & 5.6 & 18.8 & 29.4 & 68.0 & 121.0 &30 &  & 1.5 & 5.7 & 9.5 & 36.9 & 53.7 &188\\
DE++ \cite{dong2021dual} &  & 8.8 & 21.9 & 30.2 & 67.4 & 128.3 &38 &  & 5.3 & 18.4 & 29.2 & 68.0 & 121.0 &31 &  & 1.7 & 5.6 & 9.6 & 37.1 & 54.1 &186\\
RIVRL 
\cite{dong2022reading} &  & 9.4 & 23.4 & 32.2 & 70.6 & 135.6 &33 &  & 5.2 & 18.0 & 28.2 & 66.4 & 117.8 &28 &  & 1.6 & 5.6 & 9.4 & 37.7 & 54.3 &183\\
XML \cite{lei2020tvr} &  & 10.0 & 26.5 & 37.3 & 81.3 & 155.1 &21 &  & 5.3 & 19.4 & 30.6 & 73.1 & 128.4 &26 &  & 1.6 & 6.0 & 10.1 & 46.9 & 64.6 &117\\
ReLoCLNet \cite{zhang2021video} &  & 10.7 & 28.1 & 38.1 & 80.3 & 157.1 &20 &  & 5.7 & 18.9 & 30.0 & 72.0 & 126.6 &26 &  & 1.2 & 5.4 & 10.0 & 45.6 & 62.3 &118 \\
JSG \cite{chen2023joint} &  & 11.3 & 29.1 & 39.6 & 80.9 & 161.0 &19 &  & 6.7 & 22.5 & 34.8 & 76.2 & 140.3 &23 &  & 1.8 & 7.2 & 11.9 & 48.3 & 69.2 &\textbf{108} \\
MS-SL++ &  & \textbf{13.6} & \textbf{33.1} & \textbf{44.2} & \textbf{83.5} & \textbf{174.5} & \textbf{14} &  & \textbf{7.0} & \textbf{23.1} & \textbf{35.2} & \textbf{75.8} & \textbf{141.1} & \textbf{22}&  & \textbf{1.8} & \textbf{7.6} & \textbf{12.0} & \textbf{48.4} & \textbf{69.7} & \textbf{108}\\ 
\cline{1-1} \cline{3-8} \cline{10-15} \cline{17-22} 
\textbf{CLIP-based features:} &  &  &  &  &  &  &  &  &  &  &  &  &  & \textbf{} & \textbf{} & \textbf{} & \textbf{} & \textbf{} \\
CLIP4Clip-meanP \cite{luo2022clip4clip} &  & 5.5 & 13.8 & 19.4 & 52.3 & 91.0 &89 &  & 11.0 & 26.4 & 36.6 & 72.0 & 146.0 &20 &  & 0.9 & 3.9 & 6.1 & 26.4 & 37.3 &565 \\
CLIP4Clip-seqLSTM \cite{luo2022clip4clip} &  & 11.0 & 25.3 & 34.0 & 73.8 & 144.1 &27 &  & 12.8 & 31.9 & 43.1 & 79.2 & 167.0 &16 &  & 1.2 & 4.0 & 6.9 & 32.0 & 44.1 &235 \\
TC-MGC \cite{jing2025tc} &  & 11.5 & 28.1 & 38.5 & 79.9 & 158.0 &21 &  & 10.1 & 28.3& 40.6 & 78.6 & 157.5 &17 &  & 0.7 & 3.4 & 6.6 & 31.5 & 42.2 &226 \\
DiCoSA \cite{ijcai2023p0104} &  & 12.1 & 28.9 & 38.9 & 78.4 & 158.4 &21 &  & 10.4 & 29.0& 40.4 & 77.7 & 157.5 &17 &  & 0.4 & 2.7 & 5.1 & 29.7 & 37.9 &246 \\
CLIP4Clip-seqTransf \cite{luo2022clip4clip} &  & 12.7 & 30.1 & 39.8 & 79.2 & 161.8 &19 &  & 12.7 & 30.1 & 43.8 & 79.8 & 168.5 &16 &  & 1.1 & 4.2 & 7.3 & 32.8 & 45.4 &231 \\
X-Pool \cite{gorti2022x} &  & 13.5 & 31.3 & 40.3 & 77.8 & 162.9 &19 &  & \textbf{13.1} & \textbf{33.8} & 45.5 & 80.5 & 172.9 &13 &  & 1.2 & 4.3 & 7.6 & 35.5 & 48.7 &190 \\
T-MASS \cite{wang2024text} &  & 14.1 & 32.6 & 42.8 & 79.1 & 168.6 &17 &  & 12.8& 31.6 & 44.1 & 79.9 &168.4  & 15 & &0.9 & 3.8 & 6.6 & 31.5 & 47.4 &231 \\
MeVTR\cite{zhang2023multi}
 &  & 13.1 & 32.0 & 40.6 & 79.5 & 165.2 &18 &  & 13.0 & 32.3 & 44.5 & 80.8 & 170.6 &14 &  & 1.3 & 4.5 & 8.0 & 36.7 & 50.5 &185 \\
DL-DKD \cite{dong2023dual} &  & 14.4 & 34.9 & 45.8 & 84.9 & 179.9 &13 &  & 8.0 & 25.0 & 37.5 & 77.1 & 147.6 &19 &  & - & - & - & - & - \\
UCoFiA \cite{wang2023unified} &  & 15.0 & 35.3 & 46.2 & 84.6 & 181.1 &14 &  & 11.6 & 30.1& 42.3 & 79.5 & 163.5 &16 &  & 0.9 & 3.8 & 6.3 & 31.0 & 41.1 &234 \\
X-CLIP \cite{ma2022x} &  & 15.5 & 35.8 & 46.7 & 85.0 & 183.1 &13 &  & 10.7 & 29.3& 41.9 & 78.8 & 160.8 &16 &  & 0.9 & 3.4 & 6.3 & 31.7 & 42.4 &232 \\
MS-SL++ &  & \textbf{23.4} & \textbf{47.2} & \textbf{58.3} & \textbf{90.1} & \textbf{219.0} &\textbf{6} &  & 12.7 & 33.1 & \textbf{46.0} & 82.1 & \textbf{173.9} &\textbf{13} &  & \textbf{1.5} & \textbf{5.2} & \textbf{9.3} & \textbf{39.2} & \textbf{55.2} &\textbf{171} \\ 
\bottomrule
\end{tabular}
}
\end{table*}

\subsection{Comparison with the State-of-the-art} \label{ssec:exp-sota}

To validate the effectiveness of our method on PRVR task, we conduct performance comparisons on conventional CNN-based features and CLIP-based features. For the CNN-based features, as there are few models specifically designed for PRVR, in addition to comparing with existing method JSG \cite{chen2023joint}, we additionally introduce models targeted at current T2VR, \ie 
VSE++~\cite{faghri2017vse++}, W2VV~\cite{dong2018predicting}, CE~\cite{liu2019use}, W2VV++~\cite{li2019w2vv++}, DE~\cite{dong2019dual}, HTM \cite{miech2019howto100m}, HGR~\cite{chen2020fine}, DE++~\cite{dong2021dual} and models developed for VCMR, \ie XML \cite{lei2020tvr} and ReLoCLNet~\cite{zhang2021video}. For the CLIP-based features, we compare  our method with several recent CLIP-based T2VR models, including  CLIP4Clip~\cite{luo2022clip4clip}, X-Pool\cite{gorti2022x}, X-CLIP\cite{ma2022x}, MeVTR\cite{zhang2023multi} UCoFiA\cite{wang2023unified}, DiCoSA \cite{ijcai2023p0104}, T-MASS\cite{wang2024text} and TC-MGC\cite{jing2025tc}, as well as DL-DKD~\cite{dong2023dual}, which employs CLIP as a teacher model to perform distillation learning for PRVR.
All VCMR methods are two-stage, where a first-stage module is used to retrieve candidate videos followed by a second-stage module to localize specific moments in the candidate videos. As moment annotations are unavailable for PRVR, we have re-trained them (with their moment localization modules removed) using the same video features as ours.

\begin{figure*}[tb!]
\centering\includegraphics[width=1.9\columnwidth]{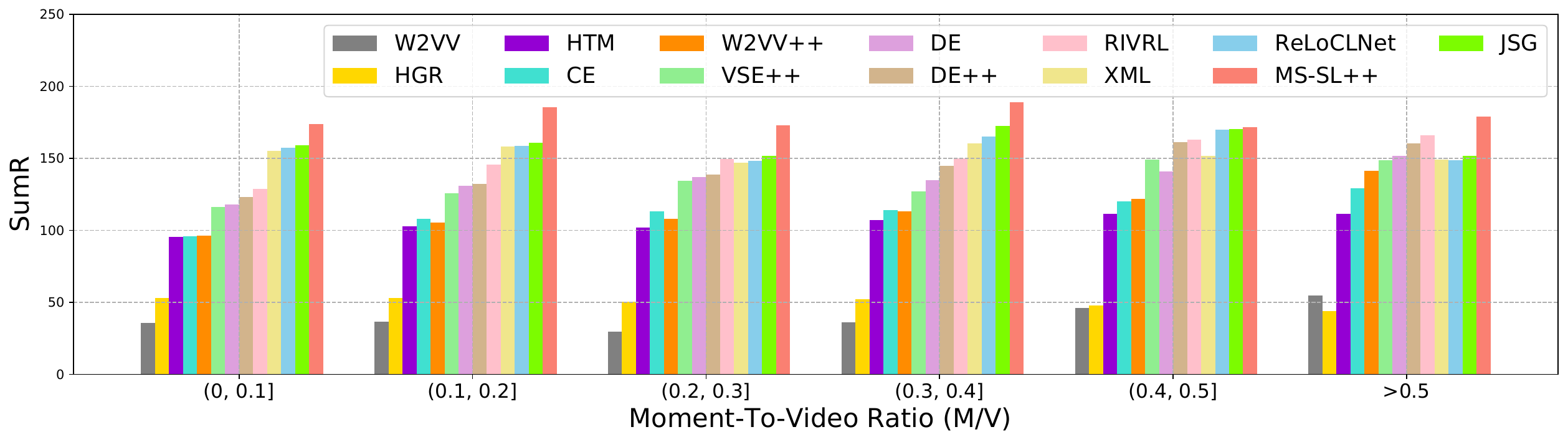}
\caption{Performance of different models on different types of queries.Queries are grouped according to their M/V. The minimal performance is in the group of (0, 0.2] where M/V is the smallest, showing video retrieval for this group is more challenging. 
Observing the figure from left to right, the average performance of the twelve compared models increases along with query's M/V, from 111.2, 118.1, 117.5, 123.1, 129.5 and 129.7, showing the current video retrieval baseline models better address queries of larger M/V.} 
\label{fig:group_recalls}
\end{figure*}

\subsubsection{Overall performance comparison}
\label{sec:perform_compare}
Table \ref{tab:sota} summarizes the performance comparison on three PRVR benchmarks. 
In the case of employing CNN-based visual features, our proposed model consistently outperforms all  T2VR models with a clear margin. Even the best performing model RIVRL among the T2VR models, our model outperforms it by 38.9 in terms of SumR.
As these models focus on the whole similarity between videos and queries, the results allow us to conclude that such similarity modeling is suboptimal for PRVR.
For the XML and ReLoCLNet models solving VCMR task, they perform better than the T2VR models, but they are still worse than ours. 
They focus on retrieving moments, which to some extent model the partial relevance, but they compute the similarity only in terms of a specific scale. While the PRVR-specific model JSG achieves comparable results to our model on the ActivityNet-Captions and Charades-STA datasets, our MS-SL++ outperforms it with a clear margin on the TVR dataset.
We attribute the good performance of our proposed model to its joint consideration of  similarity at both the clip and frame scales.

On both ActivityNet-Captions and Charades-STA datasets, our model still performs the best. The results again verify the effectiveness of our model for measuring partial relevance between videos and queries.
Interestingly, we observe that HTM performs badly on TVR and ActivityNet-Captions, while on Charades-STA it achieves the best SumR score among the T2VR models.
Recall that 
Charades-STA has the least training data among the three datasets. 
The extremely simple network structure of HTM becomes 
an advantage when training on small-scale data.
For our proposed model, it consistently performs the best on the three datasets of the varying number of training samples, which to some extent shows that our model is not sensitive to the scale of training data.

For the comparison of CLIP-based visual features, our MS-SL++ model is also at the leading position on three datasets. We further observe that models that perform well on conventional T2VR task, \textit{e.g.}, T-MASS, do not consistently achieves strong results on our proposed PRVR task. In contrast, methods that incorporate multi-granularity similarity modeling tend to perform relatively better, such as X-CLIP and UCoFiA. This again suggests that the PRVR task emphasizes fine-grained partial alignment between textual queries and untrimmed videos, rather than relying on global matching alone.
Additionally, in contrast to our model using CNN-based features, using CLIP features clearly boosts the retrieval performance on TVR and ActivityNet-Captions datasets, but it results in a performance decrease on Charades-STA.
We attribute this degradation to the nature of videos in Charades-STA, which primarily focuses on dynamic human actions. Consequently, on Charades-STA the CNN-based I3D feature that is pre-trained on human action recognition video datasets is superior to the CLIP feature that is trained on static images.

\subsubsection{Grouped performance comparison}
To gain a further understanding of the individual models, we define \textit{moment-to-video ratio} (M/V) per 
query, which is measured as its corresponding moment's length ratio in the entire video.
Smaller M/V indicates less relevant content while more irrelevant content \wrt the query.
In other words, the smaller M/V to some extent means a lower relevance of a query to its corresponding video, while the larger one indicates a higher relevance.
According to M/V, queries can be automatically classified into different groups, which enables a fine-grained analysis of how a specific model responds to the different types of queries.
For the TVR dataset, we divided the 10,895 test queries into six groups according to their M/V, with the performance of each group shown in Fig.\ref{fig:group_recalls}.
Unsurprisingly, our model consistently performs the best in all groups.
Observing the figure from left to right, the average performance of the twelve compared models increases along with the M/V, from 111.2, 118.1, 117.5, 123.1, 129.5 and 129.7.
The performance in the group with the lowest M/V is the smallest, while the group with the highest M/V is the largest.
The result allows us to conclude that the current video retrieval baseline models better address queries of larger relevance to the corresponding video. 
By contrast, the performance we achieved is more balanced in all groups. This result shows that our proposed model is less sensitive to irrelevant content in videos.

\subsubsection{Robustness analysis}\label{sec:robust}
\begin{figure}[tb!]
\centering\includegraphics[width=1\columnwidth]{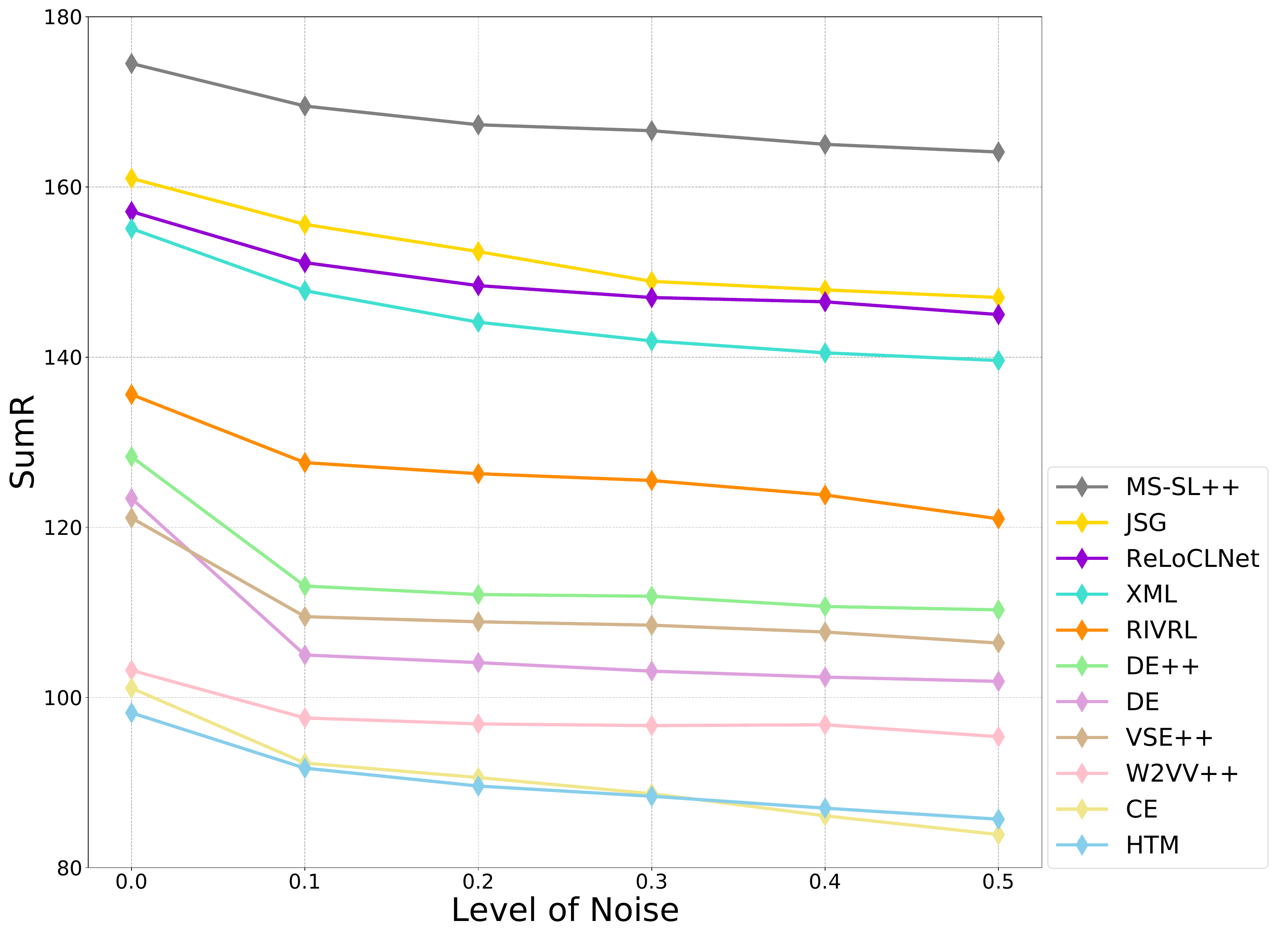}
\caption{Performance curves of different methods with respect to the artificial noise on the TVR dataset. Our proposed method consistently outperforms the others given the varied levels of noise.
}\label{fig:noise}
\end{figure}

In this section, we study the robustness of each method by adding extra noise.
Concretely, following the stability verification experiment described in \cite{yang2021deconfounded}, for each test video of $h$ seconds, we add $p$-level noise, namely a randomly-generated video clip of $h\times p$ seconds is inserted at the beginning of the video.
The frame features of the inserted video clip are randomly generated from a normal distribution. The mean and standard deviation of the normal distribution are set to the mean and standard deviation of all frame features from training videos, respectively.
The performance curves with respect to the level of noise on the TVR dataset are shown in Fig.~\ref{fig:noise}.
As expected, the performance of all methods declines as more noise is added.
Across all noise levels, our proposed method is still better than other methods.

\subsection{Generalization Ability on Unseen Data}
In order to evaluate the models' generalization ability on unseen datasets, we conduct a cross-dataset evaluation, using TVR for training and ActivityNet-Captions for testing. The CLIP-based baselines are compared. As shown in Table \ref{tab:cross dataset}, the proposed model has better generalization ability.

\begin{table}[]
\centering
\renewcommand{\arraystretch}{1.2}
\caption{Cross-dataset evaluation. Training: TVR. Test: ActivityNet-Captions.}
\label{tab:cross dataset}
\scalebox{0.8}{
\begin{tabular}{llcccccc}
\toprule
\multicolumn{2}{l}{} & \multicolumn{1}{l}{R@1} & \multicolumn{1}{l}{R@5} & \multicolumn{1}{l}{R@10} & \multicolumn{1}{l}{R@100} & \multicolumn{1}{l}{SumR} & \multicolumn{1}{l}{Med R} \\ \hline
\multicolumn{2}{l}{X-CLIP\cite{ma2022x}} & 1.9 & 5.5 & 7.9 & 26.6 & 41.9 & 384 \\
\multicolumn{2}{l}{TC-MGC\cite{jing2025tc}} & 2.2 & 8.7 & 13.9 & 44.2 & 69.1 & 142 \\
\multicolumn{2}{l}{UCoFiA\cite{wang2023unified}} & 2.9 & 9.8 & 14.8 & 45.2 & 72.7 & 134 \\
\multicolumn{2}{l}{DiCoSA\cite{ijcai2023p0104}} & 3.2 & 10.2 & 15.7 & 46.5 & 75.6 & 122 \\
\multicolumn{2}{l}{T-MASS\cite{wang2024text}} & 3.0 & 10.4 & 16.8 & 48.5 & 78.7 & 112 \\
\multicolumn{2}{l}{X-Pool\cite{gorti2022x}} & 3.1 & 10.6 & 17.2 & 48.8 & 79.6 & 109 \\
\multicolumn{2}{l}{CLIP4clip-seqTransf\cite{luo2022clip4clip}} & 3.3 & 10.9 & 17.1 & 49.0 & 80.2 & 108 \\
\multicolumn{2}{l}{MeVTR\cite{zhang2023multi}} & 3.5 & 10.8 & 17.8 & 49.3 & 81.3 & 106 \\
\multicolumn{2}{l}{MS-SL++} & \textbf{3.8} & \textbf{12.0} & \textbf{18.5} & \textbf{50.4} & \textbf{84.7} & \textbf{98} \\ \bottomrule
\end{tabular}
}
\end{table}

\begin{figure}[tb!]
\subfigure[IoU=0.5]{
\includegraphics[width=0.47\columnwidth]{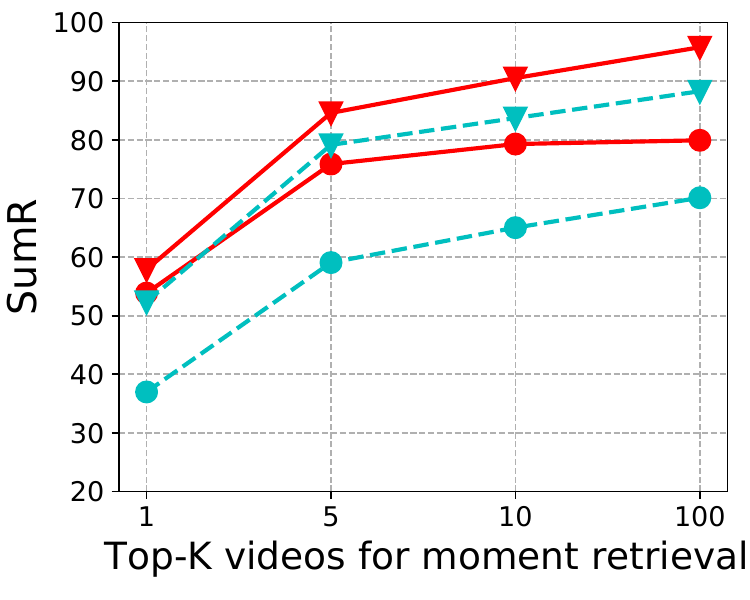}
}
\subfigure[IoU=0.7]{
\includegraphics[width=0.47\columnwidth]{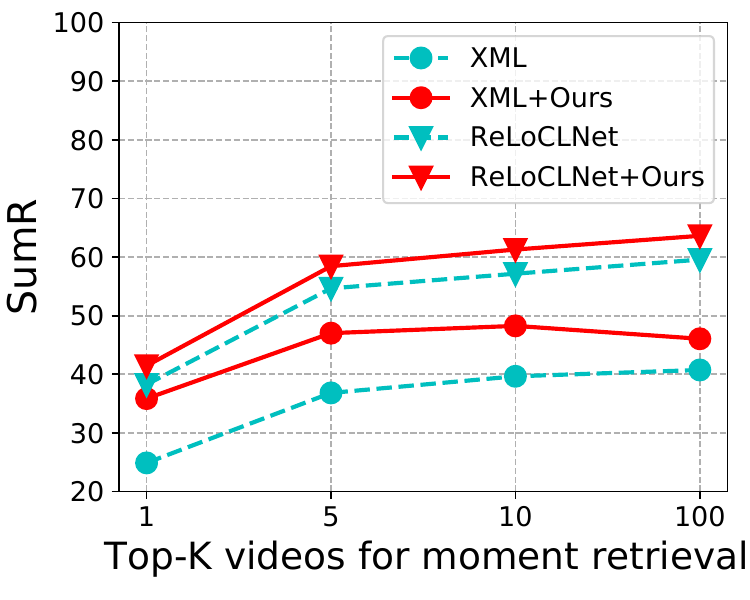}
}
\caption{Performance of XML and ReLoCLNet without/with our model as the first stage for VCMR. 
}\vspace{-4mm}\label{fig:replace_experiment}
\end{figure}

\subsection{PRVR for VCMR}\label{ssec:prvr4vcmr}
Although our model is proposed for PRVR, it can also be used for the first stage of VCMR to retrieve videos that may contain the target moment. 
In this experiment, the performance is evaluated in terms of \textit{Recall@$k$, IoU=$\mu$ ($k\in\{1,5,10,100\}, \mu\in\{0.5,0.7\}$)} that are commonly used for VCMR.
To demonstrate its potential for VCMR, we replace the first stage of two VCMR models, \ie XML \cite{lei2020tvr} and ReLoCLNet~\cite{zhang2021video}, with our model, while keeping the other processes. All the models use both visual feature and subtitle feature for video representation.
Fig.~\ref{fig:replace_experiment} shows the performance of the original models and the replaced ones on the TVR dataset.
Here, we report sum of all recalls \textit{SumR} in different IoU.
After replacing the first stage with our model, all models achieve a consistent performance gain.
This demonstrates that our method can be used for improving VCMR.

\begin{figure}[tb!]
\centering\includegraphics[width=1\columnwidth]{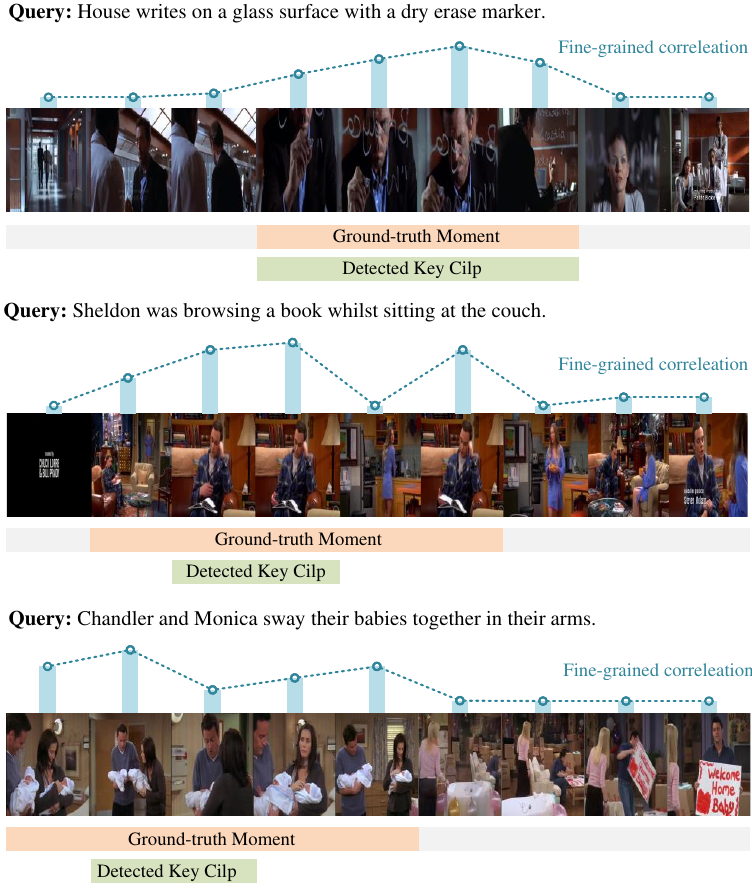}
\caption{The PRVR visualization results on TVR dataset.
}
\label{fig:visualize}
\end{figure}

\begin{figure*}[tb!]
\centering\includegraphics[width=2\columnwidth]{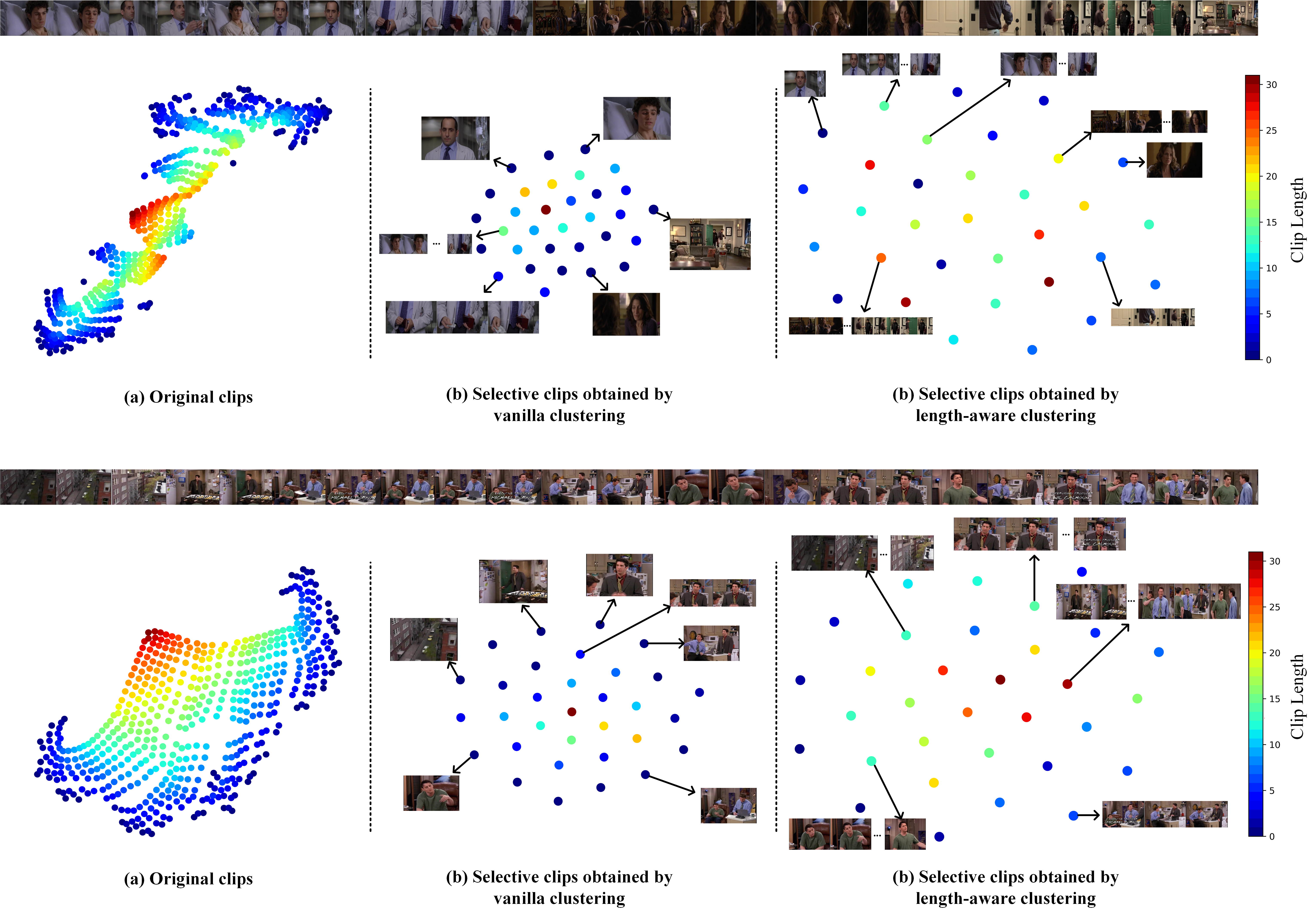}
\caption{T-SNE visualization of (a) original clips, (b) selective clips obtained by vanilla clustering, and (c) selective clips obtained by length-aware clustering, respectively. Dots with the different color indicate clips of different length, and dots that are closer together are more semantically similar. Selective clips obtained by our proposed length-aware clustering are more balanced in terms of length and exhibit greater semantic diversity.  }
\label{fig:visualize_cluster}
\end{figure*}

\subsection{Qualitative Results}
\textbf{Visualization of the PRVR results.}
To further investigate our proposed PRVR, we provide visualization examples in Fig. \ref{fig:visualize}. 
For each example, we visualize the detected key clip and frame-level fine-grained correlation.
Recall that the key clip is detected by the clip-scale SL branch, while the correlation scores are obtained by our frame-scale SL branch.
In the first example, the detected key clip by our model is well aligned with the ground-truth moment, and most frames within the ground-truth moment also exhibit high correlation scores. The results to some extent demonstrate that our model is able to identify relevant content at both the coarse clip level and the fine-grained frame level. In the remaining two examples, although the detected key clip is not perfect and only slightly overlaps with the ground-truth moment, our frame-scale SL branch compensates for this by predicting high correlation scores for the relevant frames, helping to capture the finer details of the query-related content. These results again demonstrate the effectiveness of our multi-scale modeling design.

\textbf{Visualization analysis via t-SNE. }
In order to further explore the effectiveness of our length-aware clustering, we provide the visualization of the clip features using t-SNE in Fig. \ref{fig:visualize_cluster}. For each example, we visualize the distribution of original clips, selective clips obtained by vanilla clustering, and selective clips obtained by length-aware clustering. Dots with different colors indicate clips of different length, and dots that are closer together are more semantically similar.
As shown in Fig. \ref{fig:visualize_cluster} (a), many dots are clustered closely, demonstrating redundancy of original clips and highlighting the necessity for effective clip selection.
Fig. \ref{fig:visualize_cluster} (b) and (c) show the results of selective clips obtained by vanilla clustering and length-aware clustering, respectively. Vanilla clustering reduces some redundancy but fails to fully account for the variability in clip lengths. In comparison, our proposed length-aware clustering not only addresses redundancy but also better captures the semantic differences between clips of different lengths, leading to a more balanced and representative selection of key clips.

\section{Concluding Remarks} \label{sec:conc}
In this paper, we propose a novel 
T2VR subtask termed PRVR, which is more practical for handling videos with diverse query-relevant content. Different from the conventional T2VR where the video to be retrieved is assumed to trimmed and thus fully relevant \wrt a given query,
its counterpart in PRVR is 
untrimmed and partially relevant \wrt the query. 
PRVR also differs from SVMR and VCMR, serving as a crucial intermediate step to provide coarsely-retrieved videos for the two downstream tasks. 
Investigating 
PRVR is meaningful yet challenging. 
In order to learn from video-level annotations, 
we formulate the PRVR task as a MIL problem, and accordingly develop a novel MS-SL++ model to compute the video-query relevance score at both clip and frame scales in a coarse-to-fine manner. Extensive experiments on three datasets verify the effectiveness of MS-SL++ for PRVR. The model is also shown to be beneficial for VCMR.

\section{Acknowledgements}
This work was supported by the Pioneer and Leading Goose R\&D Program of Zhejiang (No. 2024C01110, No. 2023C01212), National Natural Science Foundation of China (No. 62472385, No. 62576348), Young Elite Scientists Sponsorship Program by China Association for Science and Technology (No. 2022QNRC001), Zhejiang Provincial Natural Science Foundation (No. LZ23F020004), Fundamental Research Funds for the Provincial Universities of Zhejiang (No. FR2402ZD) and Zhejiang Provincial High-Level Talent Special Support Program.


%




\ifCLASSOPTIONcaptionsoff
  \newpage
\fi



%
\bibliographystyle{IEEEtran}
\bibliography{IEEEabrv,mm2022}

\vspace{-10mm}

\begin{IEEEbiography}
[{\includegraphics[width=1in,height=1.25in,clip,keepaspectratio]{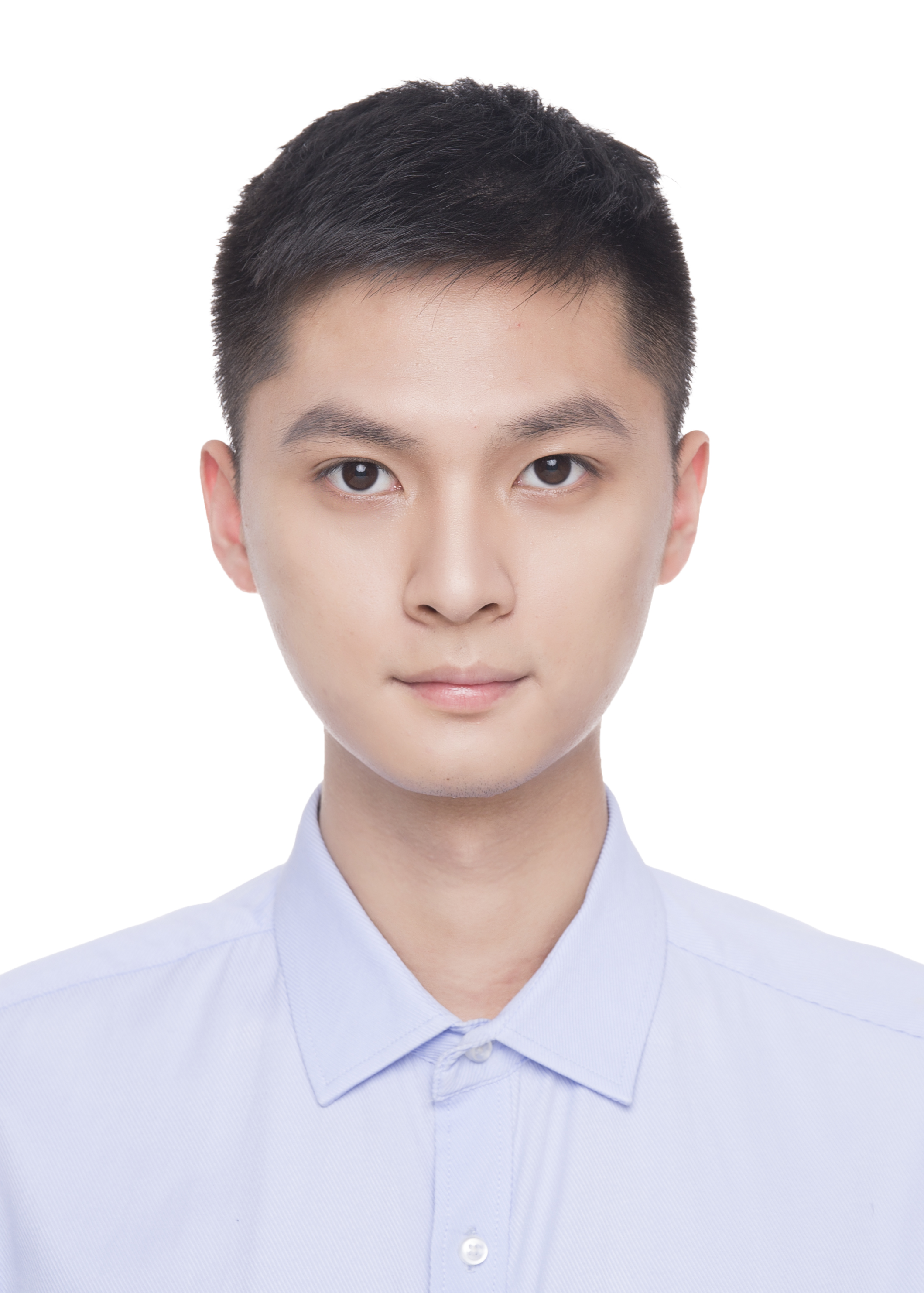}}]{Xianke Chen}
received the B.E. degree in network engineering from Zhejiang Gongshang University, Hangzhou, China, in 2020, and the M.E. degree from the College of Computer Science and Technology, Zhejiang Gongshang University, Hangzhou, China, in 2023. He is currently pursuing the Ph.D. degree with
the Big data statistics, Zhejiang Gongshang University. His research interests include multi-modal learning.
\end{IEEEbiography}

\vspace{-10mm}

\begin{IEEEbiography}
[{\includegraphics[width=1in,height=1.25in,clip,keepaspectratio]{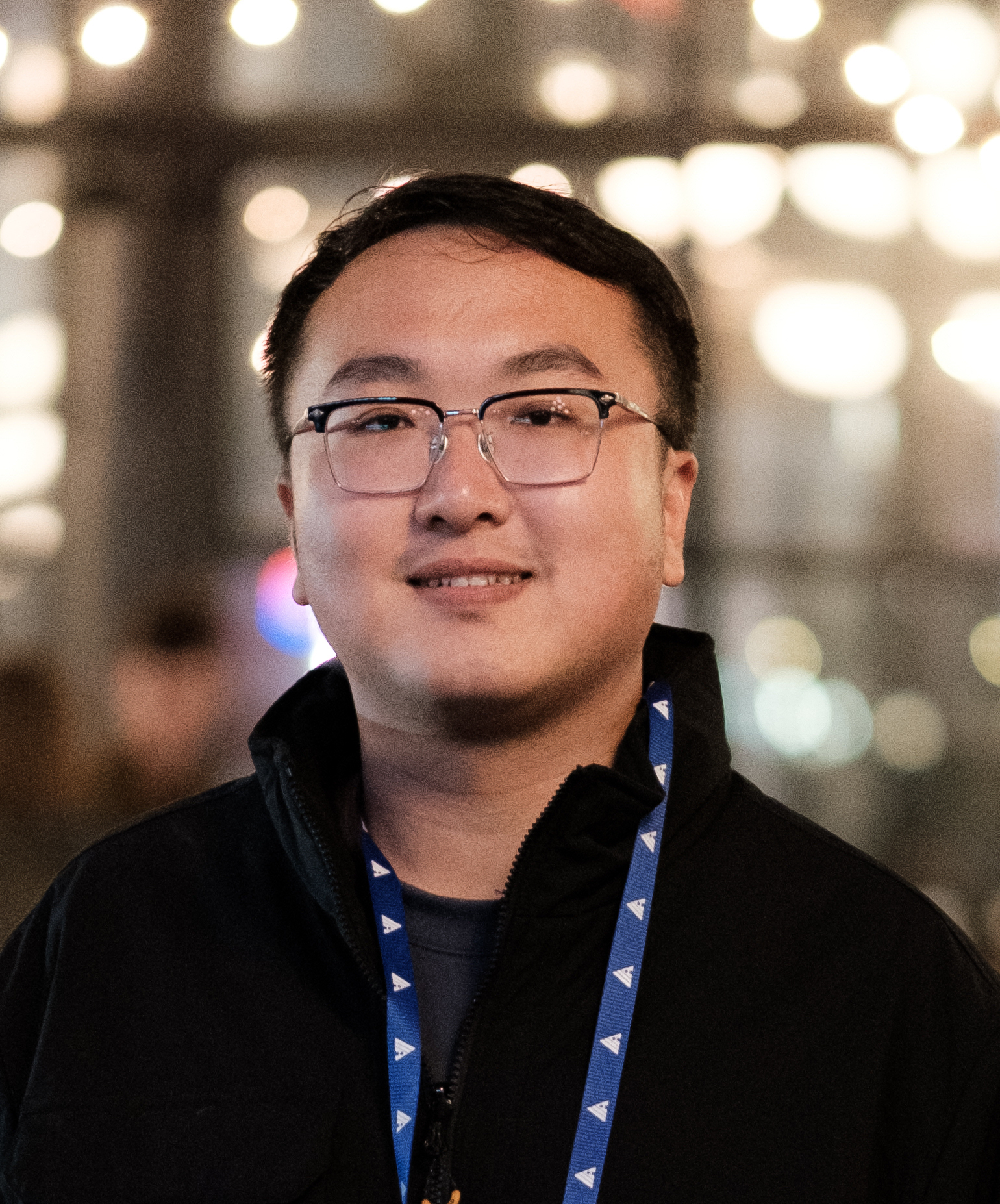}}]{Daizong Liu} received the M.S. degree in Electronic Information and Communication of Huazhong University of Science and Technology in 2021. He is currently working toward the Ph.D. degree at Wangxuan Institute of Computer Technology of Peking University. His research interests include 3D adversarial attacks, multi-modal learning, LVLM robustness, etc. He has published more than 40 papers in refereed conference proceedings and journals such as TPAMI, NeurIPS, CVPR, ICCV, ECCV, SIGIR, AAAI. He regularly serves on the program committees of top-tier AI conferences such as NeurIPS, ICML, ICLR, CVPR, ICCV and ACL.
\end{IEEEbiography}

\vspace{-10mm}

\begin{IEEEbiography}[{\includegraphics[width=1in,height=1.25in,clip,keepaspectratio]{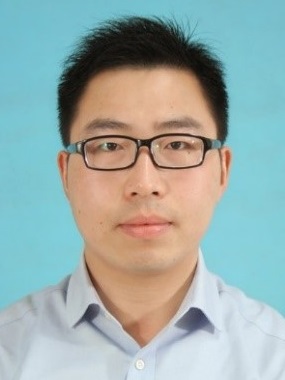}}]{Xun Yang} received the Ph.D. degree from the School of Computer and Information, Hefei University of Technology, China, in 2018. He is currently a postdoctoral research fellow with the NExT++ Research Center, National University of Singapore, Singapore. His current research interests include information retrieval, multimedia content analysis, and computer vision. He has served as the PC member and the invited reviewer for top-tier conferences and prestigious journals including ACM MM, IJCAI, AAAI, the ACM Transactions on Multimedia Computing, Communications, and Applications, IEEE Transactions on Neural Networks and Learning Systems, IEEE Transactions on Knowledge and Data Engineering, and IEEE Transactions on Circuits and Systems for Video Technology.
\end{IEEEbiography}

\vspace{-10mm}

\begin{IEEEbiography}[{\includegraphics[width=1in,height=1.25in,clip,keepaspectratio]{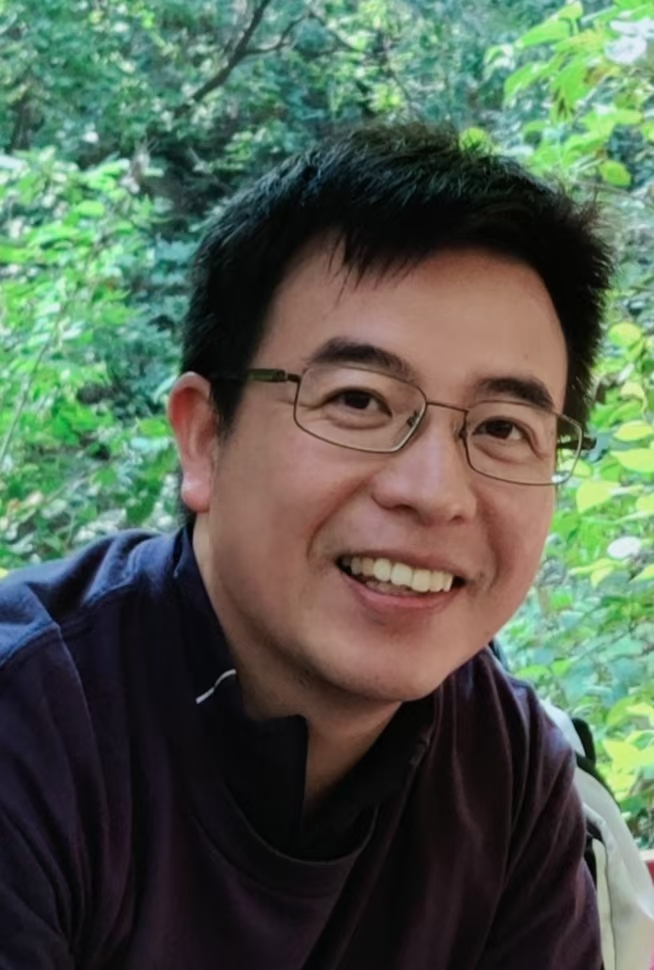}}]{Xirong Li}
(Member, IEEE) received the BS and ME degrees in computer science from Tsinghua University, Beijing, China, in 2005 and 2007, respectively, and the PhD degree in computer science from the University of Amsterdam, Amsterdam, The Netherlands, in 2012. He is currently a full professor with the School of Information, Renmin University of China, Beijing. His research focuses on multimodal intelligence. He was the recipient of CCF Science and Technology Award 2024, ACMMM 2016 Grand Challenge Award, ACM SIGMM Best PhD Thesis Award 2013, IEEE Transactions on Multimedia Prize Paper Award 2012, and the Best Paper Award of ACM CIVR 2010. He was a program co-chair for the International Conference on Multimedia Modeling 2021 and an associate editor for \emph{IET Computer Vision}. He is an associate editor for ACM TOMM and \emph{Multimedia Systems} journal.
\end{IEEEbiography}

\vspace{-10mm}

\begin{IEEEbiography}[{\includegraphics[width=1in,height=1.25in,clip,keepaspectratio]{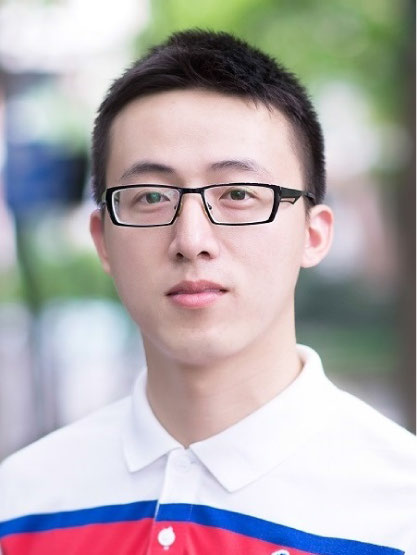}}]{Jianfeng Dong}
received the B.E. degree in software engineering from the Zhejiang University of Technology, China, in 2013, and the Ph.D. degree in computer science from Zhejiang University, China, in 2018. He is currently a Research Professor with the College of Computer Science and Technology, Zhejiang Gongshang University, Hangzhou, China. His research interests include multimedia understanding, retrieval, and recommendation. He was awarded the ACM Multimedia Grand Challenge Award and was selected into the Young Elite Scientists Sponsorship Program by the China Association for Science and Technology.
\end{IEEEbiography}

\vspace{-85mm}

\begin{IEEEbiography}[{\includegraphics[width=1in,height=1.25in,clip,keepaspectratio]{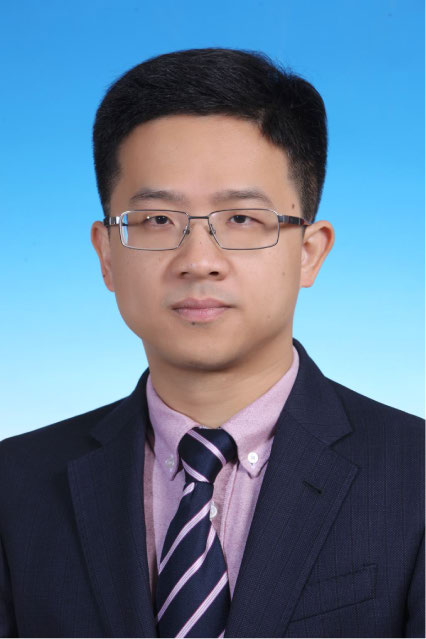}}]{Meng Wang} (Fellow, IEEE) received the B.E. and Ph.D. degrees in the special class for the Gifted Young and the Department of Electronic Engineering and Information Science, University of Science and Technology of China (USTC), Hefei, China, in 2003 and 2008, respectively. He is currently a professor with the Hefei University of Technology, China. His current research interests include multimedia content analysis, computer vision, and pattern recognition. He has authored more than 200 book chapters, journal and conference papers in these areas. He is the recipient of the ACM SIGMM Rising Star Award 2014. He is an associate editor of the IEEE Transactions on Knowledge and Data Engineering (IEEE TKDE), IEEE Transactions on Circuits and Systems for Video Technology (IEEE TCSVT), IEEE Transactions on Multimedia (IEEE TMM),and IEEE Transactions on Neural Networks and Learning Systems (IEEE TNNLS).
\end{IEEEbiography}

\vspace{-85mm}

\begin{IEEEbiography}[{\includegraphics[width=1in,height=1.25in,clip,keepaspectratio]{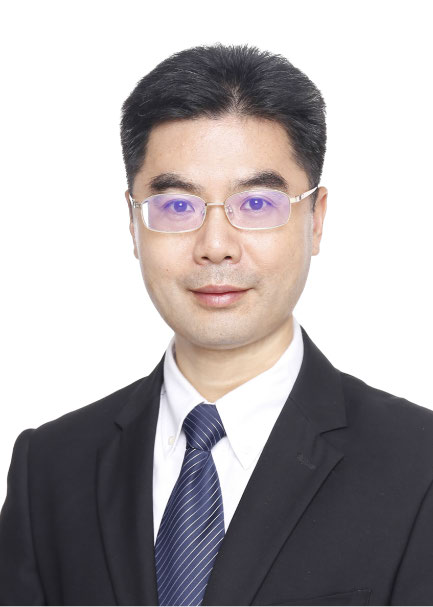}}]{Xun Wang} (Member, IEEE) received the B.S. degree in mechanics and the Ph.D. degrees in computer science from Zhejiang University, Hangzhou, China, in 1990 and 2006, respectively. He is currently a professor at the School of Computer Science and Information Engineering, Zhejiang Gongshang University, China. His current research interests include mobile graphics computing, image/video processing, pattern recognition, intelligent information processing and visualization. He is also a member of the ACM, and a senior member of the CCF.
\end{IEEEbiography}

\end{document}